\ifdefined\XeTeXrevision\else
\pdfoutput=1
\fi
\documentclass{article}
\usepackage{prl_preprint,times}

\usepackage{amsmath,amsfonts,bm}

\def\eqref#1{equation~\ref{#1}}

\def\1{\bm{1}}

\DeclareMathAlphabet{\mathsfit}{\encodingdefault}{\sfdefault}{m}{sl}
\SetMathAlphabet{\mathsfit}{bold}{\encodingdefault}{\sfdefault}{bx}{n}

\usepackage{hyperref}
\usepackage{url}
\usepackage{graphicx}
\usepackage{capt-of}
\usepackage{booktabs}
\usepackage{array}
\usepackage{amsmath}
\usepackage{amssymb}
\usepackage{multirow}
\usepackage{enumitem}
\usepackage{xcolor}
\usepackage{colortbl}
\colorlet{venuecolor}{blue!70!black}
\definecolor{oursrow}{RGB}{232,240,252}
\newcommand{\tabscale}{0.82}
\usepackage{pifont}
\usepackage{tikz}
\usetikzlibrary{arrows.meta, positioning, calc}
\usepackage{pgfplots}
\pgfplotsset{compat=1.18}
\usepgfplotslibrary{groupplots, statistics}

\definecolor{metacolor}{HTML}{0064E0}
\hypersetup{
    colorlinks=true,
    linkcolor=red,
    citecolor=metacolor,
}

\definecolor{realc}{HTML}{3A78D8}
\definecolor{synthc}{HTML}{E4573D}
\definecolor{oursc}{HTML}{2B2B2B}
\definecolor{basec}{HTML}{5C7080}
\definecolor{alertc}{HTML}{9E1F2E}
\definecolor{okc}{HTML}{3C8A4E}

\newcommand{\xstd}{96.40}
\newcommand{\xwild}{93.74}
\newcommand{\xstdgain}{4.3}
\newcommand{\xwildgain}{5.9}

\newcommand{\method}{PRL}
\DeclareMathOperator{\sg}{sg}
\DeclareMathOperator{\splus}{softplus}

\title{Relative Patch Response Learning for Generalizable AI-Generated Image Detection}

\author{
Tianyu Wang$^{1*}$,
Ouxiang Li$^{1*}$,
Yanbin Hao$^{2}$,
Zhenhua Tang$^{3}$,
Shuo Wang$^{1\dagger}$
\vspace{1mm}
\\
\small $^1$University of Science and Technology of China,
$^2$Hefei University of Technology \\
\small $^3$University of Macau \\
\small {\texttt{tianyuwang@mail.ustc.edu.cn, shuowangcv@ustc.edu.cn}}
}

\begin{document}

\maketitle
\begingroup
\renewcommand{\thefootnote}{\fnsymbol{footnote}}
\footnotetext[1]{Equal contribution.}
\footnotetext[2]{Corresponding author.}
\endgroup

\begin{abstract}
Generative models can now synthesize highly realistic images, simultaneously increasing the risks of misinformation and visual forgery. Therefore, detecting AI-generated images becomes more essential, and a reliable detector must generalize to unseen generators and stay robust to unseen perturbations in the wild. Existing detectors are typically trained on either independently collected real and generated images or aligned real-generated pairs designed to mitigate content bias. Building on aligned pairs, recent methods form a mixed view by replacing some patches of the real image with their generated counterparts. However, we find that self-attention lets real and generated patches interact, so the feature of each patch no longer reflects its own source alone. This \textbf{\textit{contextual shift}} makes a per-patch source label an imprecise target. To this end, we propose Relative Patch Response Learning (\method{}). Instead of labeling each patch, \method{} compares the same patch across two mixed views of an aligned pair and learns from its patch response, the change of its score between the views. (i)~To give precise supervision under the contextual shift, a \textbf{relative response} objective measures the responses of source-changed patches against those of source-unchanged patches, which respond to the shift alone. (ii)~To provide a reliable reference for the shift, a \textbf{reference coherence} objective keeps each group of source-unchanged patches moving as a whole. (iii)~Since the two views contain different amounts of generated content, an \textbf{area ranking} objective asks the view with the larger generated area to have a higher mean patch score. Extensive experiments demonstrate the superior performance of \method{}, which surpasses the best prior methods by \xstdgain\% and \xwildgain\% in average balanced accuracy across eight standard and three in-the-wild benchmarks, respectively. Code is available at \url{https://github.com/wittyuu/Relative-Patch-Response-Learning}.
\end{abstract}

\section{Introduction}
\label{sec:intro}

Generative models nowadays can produce highly realistic images~\citep{karras2019stylegan,dhariwal2021diffusion,rombach2022ldm,esser2024sd3,lu2023seeing,li2026easier}, expanding the possibilities for visual content creation~\citep{openai2025imagegen,labs2025kontext,wu2025qwenimage} while increasing the risks of misinformation and visual forgery~\citep{feuerriegel2023disinformation,chen2025syntheticpolitics,nightingale2022faces,yang2024fakeprofiles}. These concerns motivate AI-generated image detection, which aims to distinguish AI-generated images from real ones~\citep{wang2020cnnspot,frank2020frequency,chai2020makes,ojha2023univfd}. The task has two central challenges. The first is \textbf{\textit{generalization}}: detectors are trained on a limited set of generators but must recognize images from unseen models~\citep{corvi2023detection,zhu2023genimage}. The second is \textbf{\textit{robustness}}: images in the wild are resized, compressed and re-shared online, and a detector must remain effective after such post-processing~\citep{grommelt2024jpeg,cozzolino2024synthwildx,yan2025aide,cavia2024wildrf}.

\begin{figure}[t]
  \centering
  \includegraphics[width=0.9\textwidth]{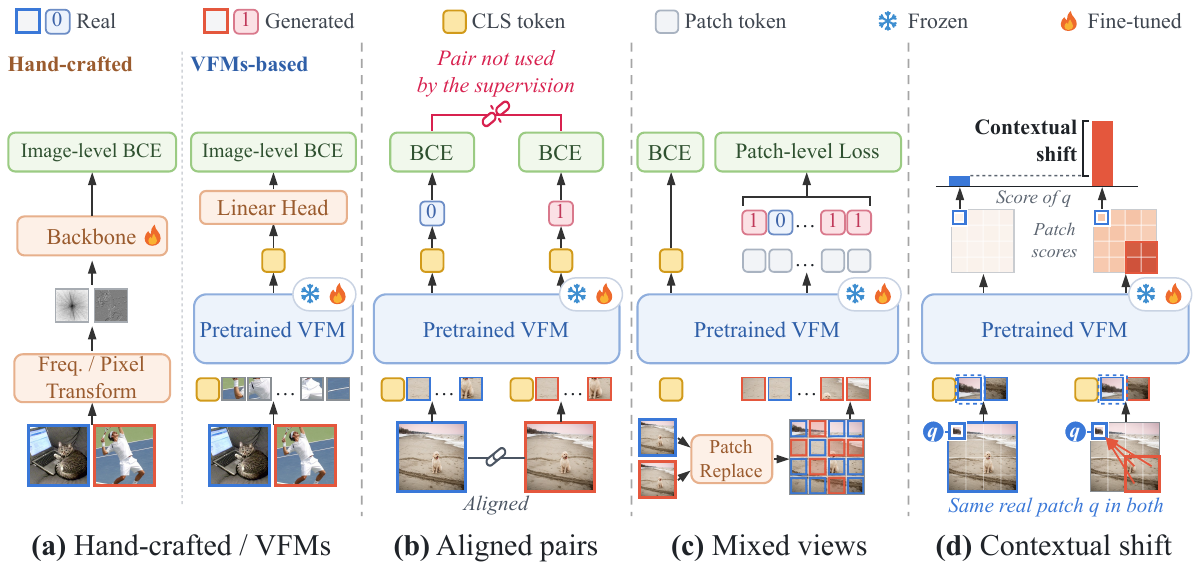}
  \vspace{-6pt}
  \caption{\textbf{Three training paradigms and the contextual shift.} (a)~Detectors built on hand-crafted features or on vision foundation models (VFMs) learn from real and generated images collected separately. (b)~Alignment-based methods pair each real image with a reconstruction of matched content, but treat the two as independent training samples with their own image-level labels; the pair itself is not used by the supervision. (c)~Patch-level methods replace some patches of a real image by their reconstruction and label each patch by its own source, in addition to the image-level label. (d)~Contextual shift: a real patch $q$ has the same input in the real image and in a mixed view, yet its score rises once other patches are replaced, while the per-patch label of (c) still marks $q$ as real.}
  \label{fig:teaser}
  \vspace{-8pt}
\end{figure}

To meet these challenges, detectors have relied on hand-crafted features that expose the traces of generation, such as frequency spectra and pixel-level residuals~\citep{frank2020frequency,tan2024npr,li2025safe}, or on the pretrained representations of vision foundation models (VFMs)~\citep{ojha2023univfd,koutlis2024rine,liu2024fatformer,yan2025effort}. In either case, the training data are real and generated images collected separately (Figure~\ref{fig:teaser}a). Such unaligned data differ in content, format and resolution as well as in source, and these differences offer a detector shortcuts around the traces of generation~\citep{guillaro2025bfree,rajan2025aligned}. Alignment-based methods mitigate content bias by pairing each real image with a reconstruction of matched content, and detectors trained on such pairs generalize better~\citep{chen2024drct,rajan2025aligned,guillaro2025bfree,chen2025dda} (Figure~\ref{fig:teaser}b). Yet the alignment only shapes the data. The two images are still trained as independent samples with image-level labels, and their correspondence is never used as supervision. Recent methods go a step further on the aligned pair. PPL~\citep{ppl2025}, for example, replaces some patches of a real image with their reconstruction, forming a mixed view in which the source of every patch is known, and supervises each patch by its own source (Figure~\ref{fig:teaser}c).

\begin{figure}[t]
  \centering
  \includegraphics[width=0.9\textwidth]{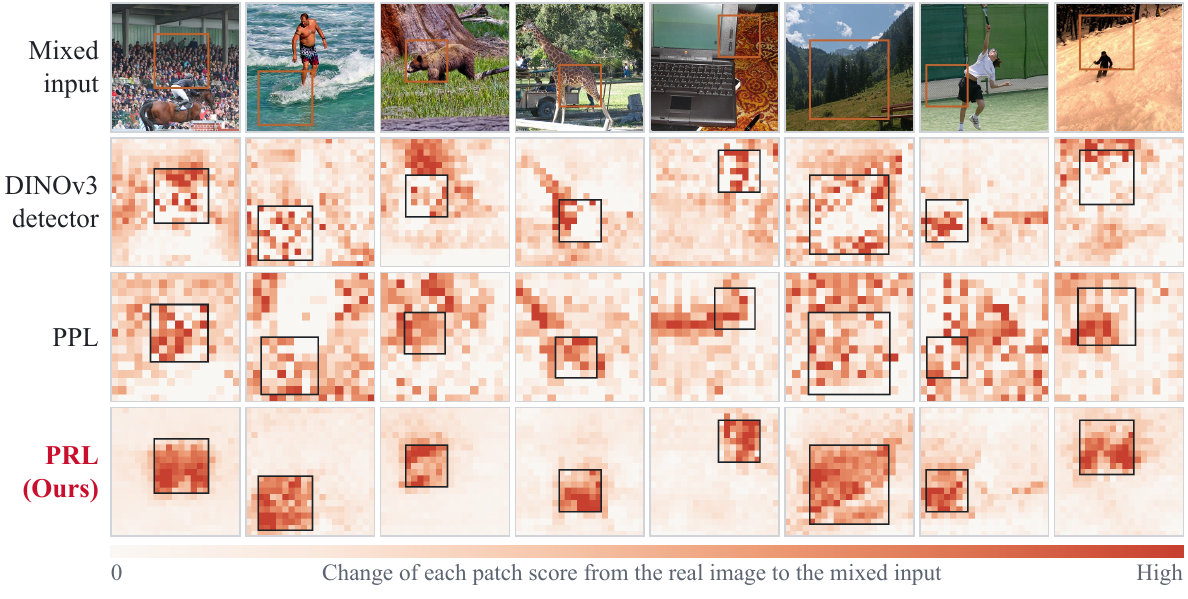}
  \vspace{-6pt}
  \caption{\textbf{Contextual shift in trained detectors.} Each mixed input takes the outlined square from the aligned reconstruction and the rest from the real image. \textbf{The maps show how much each patch score changes from the real image to the mixed input, darker for a larger change.} For DINOv3 and PPL, we train a patch head on them. Although only the square changed, the DINOv3 detector and PPL respond outside it almost as much as inside it. \method{} keeps a shift outside the square and responds inside it.}
  \label{fig:shift}
  \vspace{-8pt}
\end{figure}

However, such a per-patch label is an imprecise target. In a vision transformer, self-attention lets every token gather evidence from the whole view. In a mixed view, the traces of generation in the replaced patches therefore spread into the real ones, and the real patches in turn dilute the replaced ones, so the feature of a patch moves with its context although the patch itself does not. We call this \textbf{\textit{contextual shift}} (Figure~\ref{fig:teaser}d). The feature of a patch thus has two parts: the trace of its own source and the shift caused by the other patches, and a label that names only the source cannot tell them apart. It pushes the whole feature toward one class, so the supervision meant for the local trace is mixed with a context term, which Figure~\ref{fig:shift} makes visible in trained detectors. The label is exact about where the generated patches are, but it says nothing about what the feature of a patch should be once the rest of the view has changed.

To this end, we propose Relative Patch Response Learning (\method{}), which supervises responses instead of labels. From each aligned pair, we build two mixed views that differ in which patches are replaced (Figure~\ref{fig:method}), so that every patch has aligned content in both views and only its source may differ. A linear head maps the feature of each patch to a score, higher for stronger evidence of generation, and the change of this score between the two views is its \textbf{patch response}. For a patch whose source is the same in both views, the input is identical, so the response is the contextual shift alone; for a patch whose source differs, the response adds the effect of its own change. The same-source patches thus serve as references for the source-changed ones. \method{} has three objectives. (i)~\textbf{Relative response} supervises each source-changed group relative to its reference, which subtracts the shift and leaves a precise target. (ii)~\textbf{Reference coherence} keeps each reference group moving as a whole, which makes it a stable and reliable reference. (iii)~\textbf{Area ranking} asks the view with the larger generated area to score higher. Experiments validate this design. Supervising the fine-grained correspondence within aligned pairs while accounting for the contextual shift improves the detector on both fronts. \method{} averages \xstd\% balanced accuracy on eight standard benchmarks of diverse generators and \xwild\% on three in-the-wild benchmarks, surpassing the best previous method by \xstdgain\% and \xwildgain\%, respectively. Our contributions are as follows:

\begin{itemize}[leftmargin=*]
  \item We point out that existing methods underuse the correspondence between aligned data. Image-level methods train the two images independently, and patch-level supervision on mixed views ignores the contextual shift, which makes its per-patch labels imprecise targets.
  \item We propose \method{}, which supervises patch responses rather than labels. Relative response supervision reduces the shared contextual component against the same-source patches, reference coherence keeps the reference stable, and area-based ranking adds a view-level constraint.
  \item On eight standard benchmarks of diverse generators and three in-the-wild benchmarks, \method{} surpasses the best previous methods by \xstdgain\% and \xwildgain\% in average balanced accuracy, respectively. It also transfers to the detection of locally manipulated images with no additional training data.
\end{itemize}

\vspace{-4pt}
\section{Related Work}
\label{sec:related}

\paragraph{Forensic traces and pretrained representations.}
One family of detectors identifies generated images from the traces left by the generation process, including frequency-domain statistics~\citep{frank2020frequency,tan2024freqnet,li2025safe,karageorgiou2025spai,li2026cptformer}, pixel-level relations and residuals~\citep{tan2024npr,fu2025pid} and reconstruction or regeneration error~\citep{wang2023dire,ricker2024aeroblade,luo2024lare,yu2024semgir,chu2025fire,qi2026did}, or from the trace of the camera pipeline that generated images lack~\citep{wu2026rawrgb}. Another line builds on the pretrained representations of vision foundation models, with light classifiers on frozen features~\citep{ojha2023univfd,koutlis2024rine,yang2025d3} or with representation adaptation~\citep{liu2024fatformer,tan2025c2pclip,yan2025effort}; GAPL~\citep{qin2026gapl} organizes the feature space with generator-aware prototypes, and others scale the training set to thousands of generators~\citep{park2025community}. Both lines are typically trained on real and generated images collected separately.

\paragraph{Aligned training data.}
Separately collected real and generated images differ in content, format and resolution as well as in source, and a detector can separate them by these differences instead of by the traces of generation~\citep{guillaro2025bfree,rajan2025aligned}. Alignment removes content shortcut by giving each real image a generated counterpart of matched content. DRCT~\citep{chen2024drct} and B-Free~\citep{guillaro2025bfree} construct generated samples of similar content through reconstruction or regeneration guided by real images; AlignedForensics~\citep{rajan2025aligned} obtains a tighter spatial correspondence through VAE reconstruction; DDA~\citep{chen2025dda} further removes the high-frequency difference and interpolates pixels within each pair. Detectors trained on such pairs generalize better across generators than those trained on separately collected images~\citep{rajan2025aligned,guillaro2025bfree,chen2025dda}. In all of these methods, however, the alignment only shapes the data. The two images of a pair are trained as independent samples with image-level labels, and their correspondence is never used as supervision.

\paragraph{Mixed views and patch-level supervision.}
Mixing real and generated content yields harder training samples. Without alignment, SimLBR~\citep{dhakal2026simlbr} blends the feature of a real image with that of an unpaired generated image in the latent space of a foundation model and labels the blend as generated, which widens the training distribution but can only carry an image-level label. Within an aligned pair, the pixel-wise interpolation of DDA is again labeled at the image level, whereas PPL~\citep{ppl2025} replaces some real patches by their reconstruction and supervises each patch by its own source. Such a label, however, is decided by that patch alone and ignores the contextual shift. \method{} instead compares the same patch across two mixed views and measures its response against the patches whose source is the same in both views.

\vspace{-4pt}
\section{Method}
\label{sec:method}

\begin{figure}[t]
  \centering
  \includegraphics[width=0.9\textwidth]{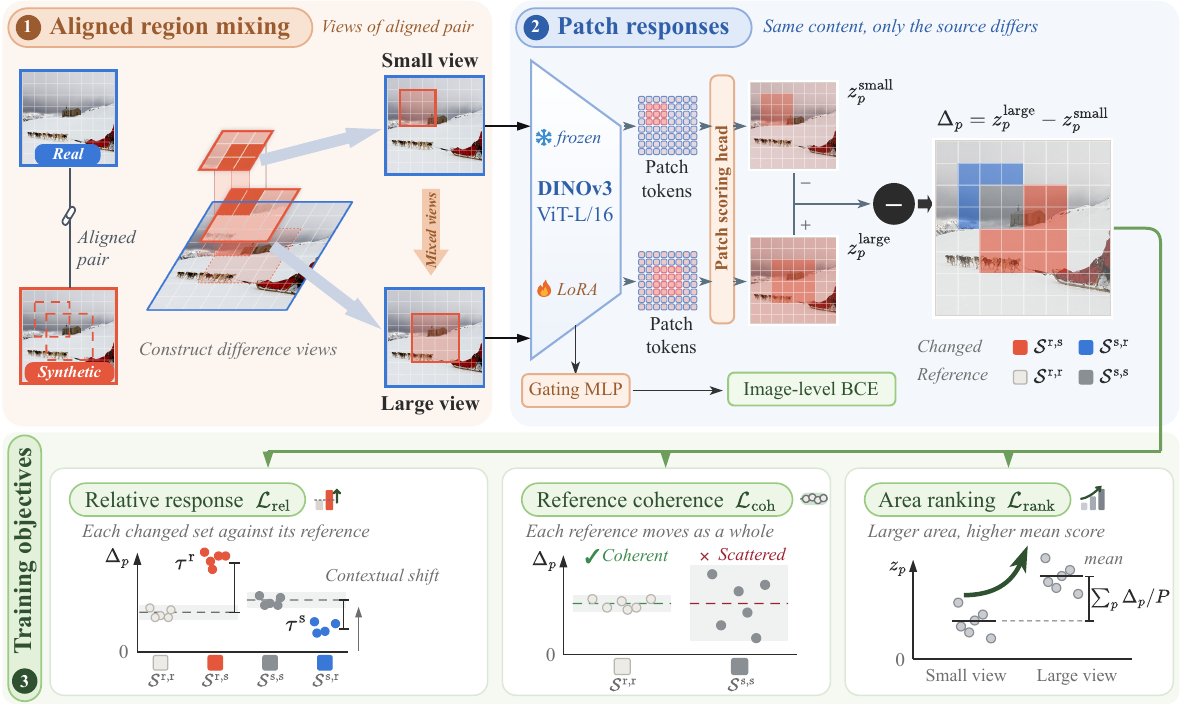}
  \vspace{-6pt}
  \caption{\textbf{Overview of \method{}.} \ding{182}~\textbf{Aligned region mixing}: two masks mix the real and the synthetic image into a small and a large view. \ding{183}~\textbf{Patch responses}: the patch scoring head scores every position in both views, and the patch response $\Delta_p$ (Eq.~(\ref{eq:response})) is read on the changed sets $\mathcal{S}^{\mathrm{r,s}}$, $\mathcal{S}^{\mathrm{s,r}}$ and on their references $\mathcal{S}^{\mathrm{r,r}}$, $\mathcal{S}^{\mathrm{s,s}}$; the gating MLP gives the image-level decision. \ding{184}~\textbf{Training objectives}: relative response, reference coherence and area ranking.}
  \label{fig:method}
  \vspace{-8pt}
\end{figure}

In this section, we detail \method{}, whose framework is shown in Figure~\ref{fig:method}. Specifically, \method{} supervises a ViT detector with the patch response of an aligned pair: how the score at each patch position changes between two mixed views of the pair. It has two parts: aligned region mixing, which constructs the mixed views, and response learning, which learns the traces of generation from the patch responses. Given a real image and its aligned synthetic image, we mix them patch by patch into two views and train the detector on how each patch score changes between the views.

\vspace{-4pt}
\subsection{Aligned region mixing}
\label{sec:views}

Given a real image $\bm{I}^{\mathrm{r}}$ and its synthetic image $\bm{I}^{\mathrm{s}}$, where the synthetic image is aligned with the real image, we use $\bm{\Omega}$ to split both images into $P$ patches as $\mathcal{X}^{\mathrm{r}}=\bm{\Omega}(\bm{I}^{\mathrm{r}})=\{\bm{x}^{\mathrm{r}}_p\}_{p=1}^{P}$ and $\mathcal{X}^{\mathrm{s}}=\bm{\Omega}(\bm{I}^{\mathrm{s}})=\{\bm{x}^{\mathrm{s}}_p\}_{p=1}^{P}$, respectively, where $\bm{\Omega}$ is the function that splits an image into non-overlapping patches on the patch grid of the encoder and $P$ is the number of patches. We then define a binary mixture mask $M=(m_1,\dots,m_P)\in\{0,1\}^{P}$ to construct a new synthetic image, which we call a mixed view, from the patches
\begin{equation}
  \hat{\bm{x}}_p = m_p\,\bm{x}^{\mathrm{s}}_p + (1-m_p)\,\bm{x}^{\mathrm{r}}_p ,
  \label{eq:view}
\end{equation}
so that the mixed view consists of the patches $\hat{\mathcal{X}}=\{\hat{\bm{x}}_p\}_{p=1}^{P}$. The mask is generated at random: $m_p=1$ takes the patch $\bm{x}^{\mathrm{s}}_p$ of $\bm{I}^{\mathrm{s}}$ at position $p$, and $m_p=0$ takes $\bm{x}^{\mathrm{r}}_p$ of $\bm{I}^{\mathrm{r}}$. Thus, different masks $M$ yield a large set of mixed views for training.

For each aligned pair, we sample two masks to construct two mixed views. The sum $\sum_p m_p$ counts the synthetic patches of a view, so a larger sum means more synthetic patches, and we require the two sums to differ; Appendix~\ref{app:impl} details how the masks are sampled. Ordered by the sum, we denote the two masks by $M^{\mathrm{small}}$ and $M^{\mathrm{large}}$, with entries $m^{\mathrm{small}}_p$ and $m^{\mathrm{large}}_p$, and the mixed views they construct by $\hat{\mathcal{X}}^{\mathrm{small}}=\{\hat{\bm{x}}^{\mathrm{small}}_p\}_{p=1}^{P}$ and $\hat{\mathcal{X}}^{\mathrm{large}}=\{\hat{\bm{x}}^{\mathrm{large}}_p\}_{p=1}^{P}$. We then define four sets of positions by the patches the two views take:
\begin{equation}
  \begin{aligned}
    \mathcal{S}^{\mathrm{r,r}} &= \{p : \hat{\bm{x}}^{\mathrm{small}}_p=\bm{x}^{\mathrm{r}}_p,\ \hat{\bm{x}}^{\mathrm{large}}_p=\bm{x}^{\mathrm{r}}_p\}, &\qquad \mathcal{S}^{\mathrm{r,s}} &= \{p : \hat{\bm{x}}^{\mathrm{small}}_p=\bm{x}^{\mathrm{r}}_p,\ \hat{\bm{x}}^{\mathrm{large}}_p=\bm{x}^{\mathrm{s}}_p\},\\
    \mathcal{S}^{\mathrm{s,s}} &= \{p : \hat{\bm{x}}^{\mathrm{small}}_p=\bm{x}^{\mathrm{s}}_p,\ \hat{\bm{x}}^{\mathrm{large}}_p=\bm{x}^{\mathrm{s}}_p\}, &\qquad \mathcal{S}^{\mathrm{s,r}} &= \{p : \hat{\bm{x}}^{\mathrm{small}}_p=\bm{x}^{\mathrm{s}}_p,\ \hat{\bm{x}}^{\mathrm{large}}_p=\bm{x}^{\mathrm{r}}_p\}.
  \end{aligned}
  \label{eq:sets}
\end{equation}
On $\mathcal{S}^{\mathrm{r,s}}$ and $\mathcal{S}^{\mathrm{s,r}}$ the patch changes source between the views; these are the changed sets in Figure~\ref{fig:method}, red and blue. On $\mathcal{S}^{\mathrm{r,r}}$ and $\mathcal{S}^{\mathrm{s,s}}$ the patch stays the same; these are the reference sets, light and dark grey. Because the pair is aligned, the patch at every position shows aligned content in both views, and only its source may differ.

\vspace{-4pt}
\subsection{Response learning}
\label{sec:response}

\paragraph{Patch response.}
Both mixed views are encoded by $\mathcal{F}_\phi$, a frozen DINOv3 ViT adapted with LoRA parameters $\phi$~\citep{hu2022lora}. The encoder takes all $P$ patches of a view and returns one final-layer token per position, $\bm{h}^{\mathrm{small}}_p=\mathcal{F}_\phi(\hat{\mathcal{X}}^{\mathrm{small}})_p$ and $\bm{h}^{\mathrm{large}}_p=\mathcal{F}_\phi(\hat{\mathcal{X}}^{\mathrm{large}})_p$. A linear \emph{patch scoring head} $\bm{w}$ maps each token to a scalar \emph{patch score}, $z^{\mathrm{small}}_p=\bm{w}^{\top}\bm{h}^{\mathrm{small}}_p$ and $z^{\mathrm{large}}_p=\bm{w}^{\top}\bm{h}^{\mathrm{large}}_p$, higher for stronger evidence of synthesis. The \emph{patch response} at position $p$ is the change of its score from the small to the large view,
\begin{equation}
  \Delta_p = z^{\mathrm{large}}_p - z^{\mathrm{small}}_p = \bm{w}^{\top}\big(\bm{h}^{\mathrm{large}}_p-\bm{h}^{\mathrm{small}}_p\big).
  \label{eq:response}
\end{equation}

Since the content at $p$ is aligned across the two views, $\Delta_p$ is not driven by content. For $p\in\mathcal{S}^{\mathrm{r,r}}\cup\mathcal{S}^{\mathrm{s,s}}$, the patch at position $p$ is identical in the two views, so $\Delta_p$ comes only from the patches that changed elsewhere in the image. This is the contextual shift (Figure~\ref{fig:shift}), and the reference sets measure it directly. For $p\in\mathcal{S}^{\mathrm{r,s}}\cup\mathcal{S}^{\mathrm{s,r}}$, $\Delta_p$ responds both to the change of the patch at $p$ and to the changed context, and a single position cannot tell the two apart. We therefore measure each changed set against the reference set with the same source in the small view: $\mathcal{S}^{\mathrm{r,s}}$ against $\mathcal{S}^{\mathrm{r,r}}$, and $\mathcal{S}^{\mathrm{s,r}}$ against $\mathcal{S}^{\mathrm{s,s}}$. The two sets of each couple start from the same source and are exposed to the same change in the rest of the image; they differ mostly in whether their own patch changes, so the mean response of the reference serves as the estimate of the contextual shift of its changed set.

\paragraph{Relative response ($\mathcal{L}_{\mathrm{rel}}$).}
A change from real to synthetic should raise the score of a position above its reference, and a change from synthetic to real should lower it below its reference. In both cases, of the two sets that share a source in the small view, the one synthetic in the large view should respond more than the one real in the large view. For the positions real and synthetic in the small view, we measure this by the \emph{relative responses}
\begin{equation}
  \tau^{\mathrm{r}} = \frac{1}{|\mathcal{S}^{\mathrm{r,s}}|}\sum_{p\in\mathcal{S}^{\mathrm{r,s}}}\Delta_p - \sg\!\Big(\frac{1}{|\mathcal{S}^{\mathrm{r,r}}|}\sum_{p\in\mathcal{S}^{\mathrm{r,r}}}\Delta_p\Big),\enspace
  \tau^{\mathrm{s}} = \sg\!\Big(\frac{1}{|\mathcal{S}^{\mathrm{s,s}}|}\sum_{p\in\mathcal{S}^{\mathrm{s,s}}}\Delta_p\Big) - \frac{1}{|\mathcal{S}^{\mathrm{s,r}}|}\sum_{p\in\mathcal{S}^{\mathrm{s,r}}}\Delta_p,
  \label{eq:relative}
\end{equation}
where $|\cdot|$ is the number of elements in a set and $\sg(\cdot)$ is the stop-gradient function: it returns its input unchanged in the forward pass and passes no gradient to it in the backward pass. Applied to the means over the references $\mathcal{S}^{\mathrm{r,r}}$ and $\mathcal{S}^{\mathrm{s,s}}$, it keeps this loss from moving the references. With the softplus activation $\splus(\cdot)=\log\big(1+\exp(\cdot)\big)$, the loss asks both relative responses to be positive,
\begin{equation}
  \mathcal{L}_{\mathrm{rel}} = \tfrac{1}{2}\big[\splus(-\tau^{\mathrm{r}}) + \splus(-\tau^{\mathrm{s}})\big].
  \label{eq:rel}
\end{equation}
If a set in $\tau^{\mathrm{r}}$ or $\tau^{\mathrm{s}}$ is empty, that term is dropped and the other is used alone. Subtracting the reference reduces the contextual shift.

\paragraph{Reference coherence ($\mathcal{L}_{\mathrm{coh}}$).}
The mean of a reference is a good estimate of the contextual shift only if its positions move together. We therefore penalize the deviation of every position in $\mathcal{S}^{\mathrm{r,r}}$ and $\mathcal{S}^{\mathrm{s,s}}$ from the mean response of its set, averaged over both sets,
\begin{equation}
  \mathcal{L}_{\mathrm{coh}} = \frac{1}{|\mathcal{S}^{\mathrm{r,r}}|+|\mathcal{S}^{\mathrm{s,s}}|}\Bigg[\sum_{p\in\mathcal{S}^{\mathrm{r,r}}}\bigg(\Delta_p-\frac{1}{|\mathcal{S}^{\mathrm{r,r}}|}\sum_{q\in\mathcal{S}^{\mathrm{r,r}}}\Delta_q\bigg)^{2}+\sum_{p\in\mathcal{S}^{\mathrm{s,s}}}\bigg(\Delta_p-\frac{1}{|\mathcal{S}^{\mathrm{s,s}}|}\sum_{q\in\mathcal{S}^{\mathrm{s,s}}}\Delta_q\bigg)^{2}\Bigg].
  \label{eq:coh}
\end{equation}
Removing the mean leaves the shift itself free: a reference may shift with the context, but it has to shift as a whole.

\paragraph{Area ranking ($\mathcal{L}_{\mathrm{rank}}$).}
The two views also differ in how much of the view is synthetic: the large view has $|\mathcal{S}^{\mathrm{r,s}}|-|\mathcal{S}^{\mathrm{s,r}}|$ more synthetic patches than the small view, a number that is always positive because the two masks have different sums. The loss
\begin{equation}
  \mathcal{L}_{\mathrm{rank}} = \frac{|\mathcal{S}^{\mathrm{r,s}}|-|\mathcal{S}^{\mathrm{s,r}}|}{P}\,\splus\!\bigg(-\frac{\sum_{p=1}^{P}\Delta_p}{|\mathcal{S}^{\mathrm{r,s}}|-|\mathcal{S}^{\mathrm{s,r}}|}\bigg)
  \label{eq:rank}
\end{equation}
asks the large view to have a higher mean patch score, that is, a positive $\sum_p\Delta_p$. Inside the softplus, the score gain is divided by the number of added synthetic patches, so the loss looks at the gain per added patch. The factor in front keeps the gradient bounded for any area gap.

\paragraph{Training and inference.}
The image-level decision comes from the \emph{detection head} $g$, a gating MLP that weights the tokens of an image, pools them and maps the result to a logit. For an image with patches $\mathcal{X}$, the detector outputs the probability $\hat{y}$ that the image is synthetic, and is trained with the binary cross-entropy $\mathcal{L}_{\mathrm{img}}$,
\begin{equation}
  \hat{y} = \sigma\big(g(\mathcal{F}_\phi(\mathcal{X}))\big), \qquad
  \mathcal{L}_{\mathrm{img}} = -y\log\hat{y}-(1-y)\log(1-\hat{y}),
  \label{eq:img}
\end{equation}
where $\sigma$ is the sigmoid function and the label $y$ is $0$ for the real image and $1$ for every image that contains synthetic patches; $\mathcal{L}_{\mathrm{img}}$ is averaged over a batch with real and synthetic images weighted equally. The training loss is
\begin{equation}
  \mathcal{L} = \mathcal{L}_{\mathrm{img}} + \lambda_{\mathrm{rel}}\,\mathcal{L}_{\mathrm{rel}} + \lambda_{\mathrm{coh}}\,\mathcal{L}_{\mathrm{coh}} + \lambda_{\mathrm{rank}}\,\mathcal{L}_{\mathrm{rank}},
  \label{eq:total}
\end{equation}
with which we train $\mathcal{F}_\phi$, $\bm{w}$ and $g$. The response terms reach the encoder through the patch scores and thereby shape the tokens that $g$ classifies. \textbf{At inference}, the detector takes a single image $\bm{I}$ and outputs $\hat{y}$ of Eq.~(\ref{eq:img}) with $\mathcal{X}=\bm{\Omega}(\bm{I})$.

\vspace{-4pt}
\section{Experiments}
\label{sec:experiments}

\vspace{-4pt}
\subsection{Experimental Setup}
\label{sec:setup-exp}
\paragraph{Implementation Details}
We build on DINOv3 ViT-L/16~\citep{simeoni2025dinov3}, kept frozen and adapted with LoRA~\citep{hu2022lora}. The training pairs are those of DDA~\citep{chen2025dda}: real images from MSCOCO~\citep{lin2014coco} and, for each, its reconstruction through the VAE of Stable Diffusion~2.1~\citep{rombach2022ldm}. Images are $336\times336$ at training and at inference. We train with Adam~\citep{kingma2015adam} at a learning rate of $10^{-4}$ and a batch of 16 pairs per GPU on two A100 GPUs, with $(\lambda_{\mathrm{rel}},\lambda_{\mathrm{coh}},\lambda_{\mathrm{rank}})=(0.3,\,0.3,\,0.1)$. Further details are given in Appendix~\ref{app:impl}.

\paragraph{Evaluation Benchmarks}
We evaluate on eleven benchmarks, which we divide into two groups. Eight standard benchmarks are assembled from known generators: GenImage~\citep{zhu2023genimage}, AIGCDetectBenchmark (AIGCDetect)~\citep{zhong2023patchcraft}, DRCT-2M~\citep{chen2024drct}, DDA-COCO and EvalGEN~\citep{chen2025dda}, Synthbuster~\citep{bammey2023synthbuster}, ForenSynths~\citep{wang2020cnnspot} and UnivFD~\citep{ojha2023univfd}; together they cover diverse generators, including GANs, diffusion models and auto-regressive models. Three in-the-wild benchmarks collect images from the web, with unknown generators and post-processing: Chameleon~\citep{yan2025aide}, SynthWildX~\citep{cozzolino2024synthwildx} and WildRF~\citep{cavia2024wildrf}. A third group, of locally manipulated images, is introduced in Section~\ref{sec:local}. Appendix~\ref{app:datasets} describes the benchmarks in more detail.

\paragraph{Evaluation Metrics and Comparative Methods}
We report balanced accuracy, the mean of the accuracies on real and on synthetic images. Following DDA, on benchmarks whose synthetic images are stored losslessly while real ones are JPEG (GenImage, AIGCDetect, ForenSynths and UnivFD), synthetic images are JPEG-compressed at quality 96 before evaluation, so that no detector can rely on the format bias. We compare with eleven detectors: UnivFD~\citep{ojha2023univfd}, FatFormer~\citep{liu2024fatformer}, NPR~\citep{tan2024npr}, DRCT~\citep{chen2024drct}, C2P-CLIP~\citep{tan2025c2pclip}, AIDE~\citep{yan2025aide}, AlignedForensics~\citep{rajan2025aligned}, Effort~\citep{yan2025effort}, SAFE~\citep{li2025safe}, DDA~\citep{chen2025dda} and GAPL~\citep{qin2026gapl}. We evaluate each method using its official checkpoint and inference code; Appendix~\ref{app:methods} describes the methods in more detail.

\vspace{-4pt}
\subsection{Cross-Benchmark Comparison}
\label{sec:main}

\begin{table}[!t]
  \caption{\textbf{Cross-benchmark comparison.} Balanced accuracy (\%) on eight standard and three in-the-wild benchmarks; EvalGEN has no real counterpart and is reported as accuracy on its synthetic images. Every compared method uses its official checkpoint and inference code. Bold: best per column; underlined: second best.}
  \label{tab:main}
  \centering
  \renewcommand{\arraystretch}{1.25}
  \scriptsize
  \setlength{\tabcolsep}{2.5pt}
\resizebox{\ifdim\width>\textwidth\textwidth\else\width\fi}{!}{\scalebox{\tabscale}{%
  \begin{tabular}{lcccccccccccc}
    \toprule\addlinespace[4pt]
    \multirow{2}{*}{Method} & \multicolumn{8}{c}{Standard Benchmarks} & \multicolumn{3}{c}{In-the-Wild Benchmarks} & \\
    \cmidrule(lr){2-9} \cmidrule(lr){10-12} \cmidrule(lr){13-13}
    & GenImage & AIGCDetect & DRCT-2M & DDA-COCO & EvalGEN & Synthbuster & ForenSynths & UnivFD & Chameleon & SynthWildX & WildRF & \textbf{Avg.} \\
    \midrule
    UnivFD\,\raisebox{-0.25ex}{\textcolor{venuecolor}{\tiny CVPR'23}} & 64.05 & 72.50 & 59.64 & 52.32 & 15.39 & 55.31 & 77.69 & 78.95 & 50.71 & 52.27 & 55.08 & 57.63 \\
    FatFormer\,\raisebox{-0.25ex}{\textcolor{venuecolor}{\tiny CVPR'24}} & 61.03 & 74.78 & 50.82 & 51.63 & 27.17 & 53.62 & 86.13 & 82.07 & 51.11 & 53.31 & 59.62 & 59.21 \\
    NPR\,\raisebox{-0.25ex}{\textcolor{venuecolor}{\tiny CVPR'24}} & 46.74 & 47.78 & 49.62 & 49.07 & 3.43 & 41.83 & 37.88 & 41.92 & 52.26 & 49.80 & 59.34 & 43.61 \\
    DRCT\,\raisebox{-0.25ex}{\textcolor{venuecolor}{\tiny ICML'24}} & 80.42 & 67.67 & 90.41 & 66.89 & 50.93 & 73.04 & 54.86 & 64.27 & 68.93 & 74.47 & 74.22 & 69.65 \\
    C2P-CLIP\,\raisebox{-0.25ex}{\textcolor{venuecolor}{\tiny AAAI'25}} & 62.09 & 70.16 & 54.29 & 50.90 & 13.82 & 68.08 & 87.00 & 84.88 & 50.60 & 53.07 & 59.14 & 59.46 \\
    AIDE\,\raisebox{-0.25ex}{\textcolor{venuecolor}{\tiny ICLR'25}} & 60.35 & 53.98 & 58.94 & 50.09 & 13.09 & 56.87 & 42.36 & 47.05 & 57.36 & 55.25 & 62.08 & 50.67 \\
    AlignedForensics\,\raisebox{-0.25ex}{\textcolor{venuecolor}{\tiny ICLR'25}} & 77.44 & 65.63 & 95.13 & 86.32 & 61.42 & 77.29 & 51.47 & 60.47 & 68.04 & 79.21 & 82.20 & 73.15 \\
    Effort\,\raisebox{-0.25ex}{\textcolor{venuecolor}{\tiny ICML'25}} & 79.55 & 77.00 & 67.42 & 51.88 & 94.41 & 75.68 & 64.17 & 70.19 & 58.76 & 58.53 & 72.95 & 70.05 \\
    SAFE\,\raisebox{-0.25ex}{\textcolor{venuecolor}{\tiny KDD'25}} & 51.14 & 50.02 & 50.42 & 50.28 & 1.10 & 52.75 & 40.67 & 43.73 & 52.51 & 49.93 & 58.97 & 45.59 \\
    DDA\,\raisebox{-0.25ex}{\textcolor{venuecolor}{\tiny NeurIPS'25}} & 91.74 & 87.83 & \underline{98.47} & \underline{92.89} & \underline{96.86} & 85.52 & 81.41 & 82.58 & \underline{82.41} & \underline{90.87} & \underline{90.20} & \underline{89.16} \\
    GAPL\,\raisebox{-0.25ex}{\textcolor{venuecolor}{\tiny CVPR'26}} & \underline{95.56} & \underline{96.08} & 95.23 & 79.27 & 96.10 & \underline{90.63} & \underline{90.94} & \underline{92.89} & 69.30 & 86.23 & 86.01 & 88.93 \\
    \midrule
    \rowcolor{oursrow} \textbf{\method{} (Ours)} & \textbf{96.83} & \textbf{97.75} & \textbf{99.27} & \textbf{94.39} & \textbf{99.14} & \textbf{97.00} & \textbf{92.57} & \textbf{94.22} & \textbf{90.01} & \textbf{93.60} & \textbf{97.59} & \textbf{95.67} \\
    \addlinespace[4pt]\bottomrule
  \end{tabular}}}
  \vspace{-8pt}
\end{table}

\paragraph{Overall comparison on multiple benchmarks.}
Table~\ref{tab:main} compares \method{} with the eleven detectors on the eleven benchmarks. \method{} reaches an average balanced accuracy of $95.67\%$, surpassing DDA at $89.16\%$ by $6.5\%$ and GAPL at $88.93\%$ by $6.7\%$, and is the best method on every benchmark. This consistent margin supports the design of \method{}: supervising patch responses against internal references, instead of labeling patches, generalizes across every benchmark. \method{} also performs best on Chameleon, the most challenging in-the-wild benchmark, with $90.01\%$ against $82.41\%$ for the strongest prior method. The two strongest prior methods fail in different places. GAPL, whose training set spans thousands of generators, stays close to \method{} on the standard benchmarks but drops on all three in-the-wild benchmarks, to $69.30\%$ on Chameleon, and to $79.27\%$ on DDA-COCO; DDA, the best previous method, is strong on most benchmarks but falls behind on the in-the-wild benchmarks and on the GAN-based ones, with $81.41\%$ on ForenSynths. Tables~\ref{tab:univfd} and \ref{tab:wild} and Appendix~\ref{app:results} report the detailed results for every subset.

\paragraph{Comparison on standard benchmarks.}
Table~\ref{tab:univfd} gives the per-subset results on UnivFD; those of GenImage, AIGCDetect, Synthbuster, DRCT-2M, DDA-COCO and EvalGEN, and ForenSynths are in Appendix~\ref{app:results}. On ForenSynths, whose generators are GANs and other CNNs, the alignment-based methods DDA and AlignedForensics, trained on VAE reconstructions, fall to $81.41\%$ and $51.47\%$, whereas \method{}, trained on the same kind of pairs, reaches $92.57\%$; on DRCT-2M, whose sixteen subsets come from Stable Diffusion models and their predecessor LDM, \method{} reaches $99.27\%$ against $98.47\%$ for DDA. GAPL is weakest on DDA-COCO, $79.27\%$ against $94.39\%$ for \method{}. \method{} is above $90\%$ on every subset except five. On the DeepFake face swaps of ForenSynths it reaches $55.26\%$, where the CLIP-based detectors are stronger. On SITD and SAN, which hold low-light photographs and super-resolved images rather than generated ones, it reaches $84.72\%$ and $88.81\%$. On the SD3.5 and Flux reconstructions of DDA-COCO it reaches $85.17\%$ and $85.19\%$.

\begin{table}[!t]
  \caption{Comparison of balanced accuracy (\%) on the UnivFD benchmark.}
  \label{tab:univfd}
  \centering
  \renewcommand{\arraystretch}{1.25}
  \scriptsize
  \setlength{\tabcolsep}{3pt}
\resizebox{\ifdim\width>\textwidth\textwidth\else\width\fi}{!}{\scalebox{\tabscale}{%
  \begin{tabular}{lcccccccccccccccccccc}
    \toprule\addlinespace[4pt]
    Method & BigGAN & CRN & CycleGAN & DeepFake & GauGAN & IMLE & ProGAN & SAN & SITD & StarGAN & StyleGAN & DALL-E & Glide-50-27 & Glide-100-10 & Glide-100-27 & Guided & LDM-100 & LDM-200 & LDM-200-cfg & \textbf{Avg.} \\
    \midrule
    UnivFD\,\raisebox{-0.25ex}{\textcolor{venuecolor}{\tiny CVPR'23}} & 87.50 & 55.67 & 96.92 & 69.39 & 98.79 & 68.11 & \underline{99.38} & 58.22 & 62.22 & 95.12 & 79.99 & 79.55 & 77.80 & 77.40 & 77.80 & 67.15 & 91.20 & 90.75 & 67.05 & 78.95 \\
    FatFormer\,\raisebox{-0.25ex}{\textcolor{venuecolor}{\tiny CVPR'24}} & 97.90 & 69.32 & \underline{99.32} & \underline{86.03} & 99.17 & 69.45 & 99.21 & 62.10 & 80.83 & \textbf{99.75} & 87.51 & 88.40 & 70.10 & 69.60 & 70.60 & 61.45 & 88.40 & 89.45 & 70.75 & 82.07 \\
    NPR\,\raisebox{-0.25ex}{\textcolor{venuecolor}{\tiny CVPR'24}} & 43.25 & 0.00 & 59.83 & 33.30 & 35.62 & 0.00 & 50.45 & 48.40 & 22.22 & 49.47 & 50.18 & 52.30 & 50.50 & 50.85 & 51.00 & 45.75 & 51.15 & 51.10 & 51.05 & 41.92 \\
    DRCT\,\raisebox{-0.25ex}{\textcolor{venuecolor}{\tiny ICML'24}} & 52.45 & 49.80 & 48.83 & 56.38 & 49.45 & 49.26 & 53.02 & 86.76 & 54.44 & 54.38 & 54.45 & 79.05 & 60.05 & 66.45 & 63.15 & 59.00 & 94.80 & 94.80 & 94.65 & 64.27 \\
    C2P-CLIP\,\raisebox{-0.25ex}{\textcolor{venuecolor}{\tiny AAAI'25}} & 92.07 & \underline{92.75} & \textbf{99.39} & \textbf{91.85} & 98.20 & \textbf{97.68} & 97.10 & 74.20 & \textbf{87.22} & \underline{99.70} & 76.97 & 84.90 & 78.90 & 78.60 & 76.35 & 57.05 & 82.70 & 83.65 & 63.40 & 84.88 \\
    AIDE\,\raisebox{-0.25ex}{\textcolor{venuecolor}{\tiny ICLR'25}} & 43.68 & 19.96 & 51.73 & 45.54 & 39.73 & 19.38 & 49.31 & 51.37 & 30.56 & 51.43 & 49.57 & 51.80 & 51.25 & 51.55 & 51.50 & 50.25 & 60.80 & 62.25 & 62.35 & 47.05 \\
    AlignedForensics\,\raisebox{-0.25ex}{\textcolor{venuecolor}{\tiny ICLR'25}} & 50.52 & 49.95 & 49.69 & 52.69 & 50.24 & 49.91 & 50.35 & 61.87 & 50.56 & 51.50 & 51.46 & 74.25 & 52.20 & 52.30 & 52.10 & 50.20 & \textbf{99.70} & \textbf{99.70} & \textbf{99.70} & 60.47 \\
    Effort\,\raisebox{-0.25ex}{\textcolor{venuecolor}{\tiny ICML'25}} & 73.60 & 39.20 & 81.49 & 64.59 & 62.52 & 29.54 & 70.08 & 58.90 & 83.89 & 96.02 & 56.42 & 78.90 & 80.70 & 82.10 & 83.45 & 59.65 & 80.15 & 79.75 & 72.65 & 70.19 \\
    SAFE\,\raisebox{-0.25ex}{\textcolor{venuecolor}{\tiny KDD'25}} & 48.95 & 0.11 & 52.58 & 49.09 & 42.60 & 0.10 & 49.96 & 50.46 & 35.83 & 49.95 & 49.99 & 49.55 & 50.90 & 51.45 & 51.15 & 50.40 & 49.15 & 49.20 & 49.45 & 43.73 \\
    DDA\,\raisebox{-0.25ex}{\textcolor{venuecolor}{\tiny NeurIPS'25}} & 90.97 & 86.98 & 72.47 & 76.53 & 92.74 & 89.71 & 92.80 & \underline{94.75} & 58.61 & 72.74 & 87.80 & 75.05 & 76.75 & 80.05 & 78.10 & 90.25 & 84.15 & 84.20 & 84.45 & 82.58 \\
    GAPL\,\raisebox{-0.25ex}{\textcolor{venuecolor}{\tiny CVPR'26}} & \underline{98.62} & 67.85 & 97.54 & 84.88 & \textbf{99.63} & 67.12 & \textbf{99.83} & \textbf{96.58} & 82.50 & 97.05 & \underline{97.89} & \underline{97.15} & \textbf{97.45} & \textbf{97.15} & \textbf{97.00} & \underline{94.15} & 97.50 & 97.55 & 97.55 & \underline{92.89} \\
    \midrule
    \rowcolor{oursrow} \textbf{\method{} (Ours)} & \textbf{99.28} & \textbf{96.07} & 97.37 & 55.26 & \underline{99.55} & \underline{96.08} & 98.99 & 88.81 & \underline{84.72} & 99.25 & \textbf{98.21} & \textbf{98.25} & \underline{96.65} & \underline{95.95} & \underline{96.25} & \textbf{94.95} & \underline{98.25} & \underline{98.25} & \underline{98.10} & \textbf{94.22} \\
    \addlinespace[4pt]\bottomrule
  \end{tabular}}}
  \vspace{-8pt}
\end{table}

\paragraph{Comparison on in-the-wild benchmarks.}
Table~\ref{tab:wild} covers images collected from the web, with unknown generators and post-processing. \method{} is above $90\%$ on all three benchmarks and on every one of their subsets; the gap to the best prior method is largest on Chameleon, where every method drops the most.

\begin{table}[!t]
  \renewcommand{\arraystretch}{1.25}
  \scriptsize
  \setlength{\tabcolsep}{3pt}
  \begin{minipage}[b]{0.632\textwidth}
    \captionof{table}{Comparison of balanced accuracy (\%) on the in-the-wild benchmarks Chameleon, SynthWildX and WildRF.}
    \label{tab:wild}
    \centering
    \resizebox{\ifdim\width>\linewidth\linewidth\else\width\fi}{!}{\scalebox{\tabscale}{%
\begin{tabular}[b]{lccccccccc}
    \toprule\addlinespace[4pt]
    Method & \multicolumn{1}{c}{Chameleon} & \multicolumn{4}{c}{SynthWildX} & \multicolumn{4}{c}{WildRF} \\
    \cmidrule(lr){2-2} \cmidrule(lr){3-6} \cmidrule(lr){7-10}
    & Chameleon & DALL-E~3 & Firefly & Midjourney & \textbf{Avg.} & Facebook & Reddit & Twitter & \textbf{Avg.} \\
    \midrule
    UnivFD\,\raisebox{-0.25ex}{\textcolor{venuecolor}{\tiny CVPR'23}} & 50.71 & 45.42 & 65.24 & 46.14 & 52.27 & 49.06 & 60.20 & 55.98 & 55.08 \\
    FatFormer\,\raisebox{-0.25ex}{\textcolor{venuecolor}{\tiny CVPR'24}} & 51.11 & 48.01 & 61.62 & 50.30 & 53.31 & 54.37 & 69.67 & 54.82 & 59.62 \\
    NPR\,\raisebox{-0.25ex}{\textcolor{venuecolor}{\tiny CVPR'24}} & 52.26 & 49.53 & 50.55 & 49.31 & 49.80 & 50.31 & 74.93 & 52.78 & 59.34 \\
    DRCT\,\raisebox{-0.25ex}{\textcolor{venuecolor}{\tiny ICML'24}} & 68.93 & 84.72 & 59.24 & 79.45 & 74.47 & 79.06 & 66.27 & 77.33 & 74.22 \\
    C2P-CLIP\,\raisebox{-0.25ex}{\textcolor{venuecolor}{\tiny AAAI'25}} & 50.60 & 49.97 & 59.73 & 49.53 & 53.07 & 51.88 & 71.60 & 53.94 & 59.14 \\
    AIDE\,\raisebox{-0.25ex}{\textcolor{venuecolor}{\tiny ICLR'25}} & 57.36 & 63.47 & 50.60 & 51.67 & 55.25 & 56.88 & 72.07 & 57.31 & 62.08 \\
    AlignedForensics\,\raisebox{-0.25ex}{\textcolor{venuecolor}{\tiny ICLR'25}} & 68.04 & 86.85 & 56.95 & \underline{93.83} & 79.21 & \underline{93.44} & 70.87 & 82.31 & 82.20 \\
    Effort\,\raisebox{-0.25ex}{\textcolor{venuecolor}{\tiny ICML'25}} & 58.76 & 59.32 & 61.27 & 55.02 & 58.53 & 74.06 & 79.33 & 65.47 & 72.95 \\
    SAFE\,\raisebox{-0.25ex}{\textcolor{venuecolor}{\tiny KDD'25}} & 52.51 & 49.65 & 50.48 & 49.65 & 49.93 & 50.94 & 74.07 & 51.90 & 58.97 \\
    DDA\,\raisebox{-0.25ex}{\textcolor{venuecolor}{\tiny NeurIPS'25}} & \underline{82.41} & \underline{92.22} & \underline{87.30} & 93.09 & \underline{90.87} & 93.12 & \underline{86.40} & \underline{91.07} & \underline{90.20} \\
    GAPL\,\raisebox{-0.25ex}{\textcolor{venuecolor}{\tiny CVPR'26}} & 69.30 & 87.06 & 87.22 & 84.39 & 86.23 & 88.44 & 82.20 & 87.40 & 86.01 \\
    \midrule
    \rowcolor{oursrow} \textbf{\method{} (Ours)} & \textbf{90.01} & \textbf{94.92} & \textbf{91.31} & \textbf{94.58} & \textbf{93.60} & \textbf{97.19} & \textbf{97.20} & \textbf{98.39} & \textbf{97.59} \\
    \addlinespace[4pt]\bottomrule
  \end{tabular}}}
  \end{minipage}\hfill
  \begin{minipage}[b]{0.355\textwidth}
    \captionof{table}{Comparison of balanced accuracy (\%) on AutoSplice at JPEG quality 100, 90 and 75.}
    \label{tab:autosplice}
    \centering
    \resizebox{\ifdim\width>\linewidth\linewidth\else\width\fi}{!}{\scalebox{\tabscale}{%
\begin{tabular}[b]{lcccc}
    \toprule\addlinespace[4pt]
    Method & JPEG 100 & JPEG 90 & JPEG 75 & \textbf{Avg.} \\
    \midrule
    UnivFD\,\raisebox{-0.25ex}{\textcolor{venuecolor}{\tiny CVPR'23}} & 64.27 & 58.12 & 57.47 & 59.95 \\
    FatFormer\,\raisebox{-0.25ex}{\textcolor{venuecolor}{\tiny CVPR'24}} & 73.78 & 57.84 & 53.88 & 61.83 \\
    NPR\,\raisebox{-0.25ex}{\textcolor{venuecolor}{\tiny CVPR'24}} & 51.37 & 50.30 & 50.18 & 50.62 \\
    DRCT\,\raisebox{-0.25ex}{\textcolor{venuecolor}{\tiny ICML'24}} & 70.04 & 62.90 & 60.84 & 64.59 \\
    C2P-CLIP\,\raisebox{-0.25ex}{\textcolor{venuecolor}{\tiny AAAI'25}} & 75.56 & 54.69 & 51.81 & 60.69 \\
    AIDE\,\raisebox{-0.25ex}{\textcolor{venuecolor}{\tiny ICLR'25}} & 55.22 & 51.52 & 51.27 & 52.67 \\
    AlignedForensics\,\raisebox{-0.25ex}{\textcolor{venuecolor}{\tiny ICLR'25}} & 50.34 & 50.12 & 50.80 & 50.42 \\
    Effort\,\raisebox{-0.25ex}{\textcolor{venuecolor}{\tiny ICML'25}} & 69.15 & 53.82 & 51.23 & 58.07 \\
    SAFE\,\raisebox{-0.25ex}{\textcolor{venuecolor}{\tiny KDD'25}} & 50.19 & 50.17 & 50.14 & 50.17 \\
    DDA\,\raisebox{-0.25ex}{\textcolor{venuecolor}{\tiny NeurIPS'25}} & 73.09 & 70.91 & 70.87 & 71.62 \\
    GAPL\,\raisebox{-0.25ex}{\textcolor{venuecolor}{\tiny CVPR'26}} & \underline{82.62} & \underline{78.91} & \underline{78.64} & \underline{80.06} \\
    \midrule
    \rowcolor{oursrow} \textbf{\method{} (Ours)} & \textbf{86.45} & \textbf{84.74} & \textbf{80.36} & \textbf{83.85} \\
    \addlinespace[4pt]\bottomrule
  \end{tabular}}}
  \end{minipage}
  \vspace{-8pt}
\end{table}

\vspace{-4pt}
\subsection{Comparison on Locally Manipulated Images}
\label{sec:local}

We additionally evaluate on two benchmarks of locally manipulated images, in which a region of a real image is replaced by generated content, AutoSplice~\citep{jia2023autosplice} and the ManipulationBench subset of DailyBench~\citep{dailybench2026}, both described in Appendix~\ref{app:datasets}.

Table~\ref{tab:autosplice} reports balanced accuracy on AutoSplice at JPEG quality $100$, $90$ and $75$. \method{} reaches $86.45\%$ at quality $100$ and remains the best at $90$ and $75$; GAPL is the strongest prior detector at $82.62\%$, DDA reaches $73.09\%$, and most of the remaining methods stay near chance. Results on the ManipulationBench subset of DailyBench are given in Appendix~\ref{app:results}, where \method{} again has the best average, $68.32\%$ against $66.88\%$ for DDA.

\vspace{-4pt}
\subsection{Evaluation on Robustness}
\label{sec:robust}

\begin{figure}[t]
  \centering
  \includegraphics[width=0.9\textwidth]{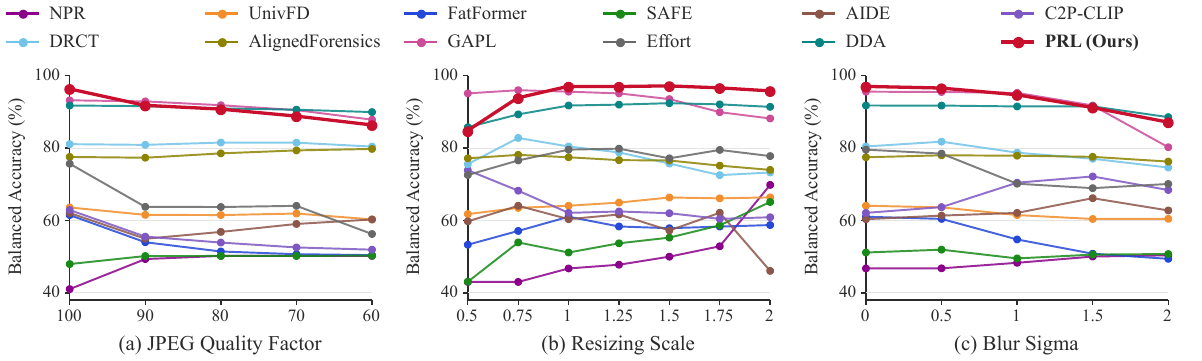}
  \vspace{-6pt}
  \caption{\textbf{Robustness to post-processing.} Balanced accuracy on debiased GenImage-JPEG96 dataset, when every test image is JPEG-compressed at decreasing quality (a), resized by a scale factor (b) or blurred with a Gaussian of increasing $\sigma$ (c).}
  \label{fig:robust}
  \vspace{-8pt}
\end{figure}

Figure~\ref{fig:robust} shows the results of three robustness evaluations on debiased GenImage-JPEG96 for all compared methods: JPEG compression at decreasing quality, resizing by a scale factor and Gaussian blur of increasing $\sigma$. \method{} keeps the highest accuracy averaged over the fourteen degraded settings, $92.1\%$ against $91.7\%$ for GAPL and $90.6\%$ for DDA. It is the best detector at every upscaling factor, staying above $95\%$ up to a scale of $2$ where GAPL falls to $88.2\%$, and it stays above $90\%$ under blur up to $\sigma=1.5$ and JPEG quality down to $80$; at $\sigma=2$ it remains within $1.6\%$ of DDA while GAPL loses $15\%$. Under the two heaviest degradations, downscaling to half the size and JPEG quality below $80$, it falls behind GAPL and DDA.

\vspace{-4pt}
\subsection{Ablation Studies}
\label{sec:ablation}

We conduct the ablation on four representative benchmarks: two standard benchmarks, GenImage and AIGCDetect, and two in-the-wild benchmarks, Chameleon and WildRF. Figure~\ref{fig:ablation} presents the results and Table~\ref{tab:ablation_full} in Appendix~\ref{app:ablation} lists the detailed numbers. Figure~\ref{fig:ablation}a compares the aligned pairs under independent image-level labels, the mixed views built from the pairs but still under the image-level label, a naive per-patch label on the mixed views, which \textbf{ignores the contextual shift}, and \method{}. The mixed views by themselves change little, $92.68\%$ against $92.65\%$ on average, and lower Chameleon by $3.1\%$. The per-patch label gains slightly on GenImage but loses a further $3.6\%$ on Chameleon. \method{} raises the average to $95.55\%$, $2.9\%$ above the image-level label and $3.7\%$ above the per-patch label. The gain thus comes from how the mixed views are supervised, not from the views themselves, and a label per patch does worse in the wild than no patch label at all.

Figure~\ref{fig:ablation}b removes one objective at a time. Each removal costs between $1.4\%$ and $2.5\%$ on average and between $3.2\%$ and $6.8\%$ on Chameleon, $\mathcal{L}_{\mathrm{rel}}$ the most. The sensitivity to the three loss weights is reported in Appendix~\ref{app:ablation}.

Besides, we adapt DDA, the best previous method, to our backbone: starting from its official code, we replace its DINOv2 backbone by DINOv3 with the same LoRA adaptation and keep the rest of its pipeline. Table~\ref{tab:dda_backbone} in Appendix~\ref{app:ablation} shows the results. The backbone lifts DDA from $89.16\%$ to $93.20\%$ on average, while \method{} on the same backbone reaches $95.67\%$.

\begin{figure}[t]
  \centering
  \includegraphics[width=\textwidth]{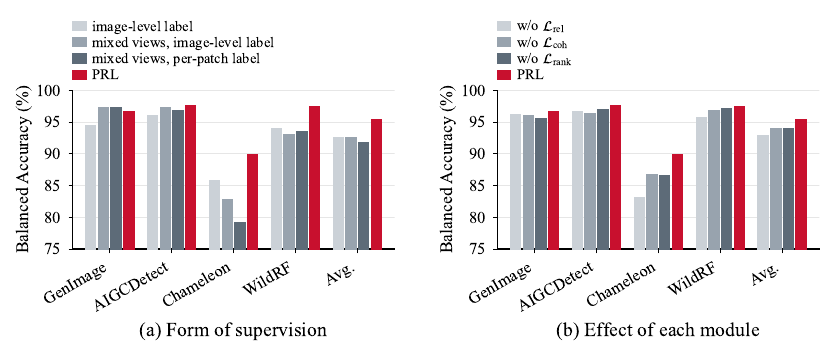}
  \vspace{-6pt}
  \caption{\textbf{Ablation.} Balanced accuracy (\%) on GenImage, AIGCDetect, Chameleon and WildRF. (a) Three ways of using the aligned pairs: independent image-level label without and with the mixed views, the per-patch label on the mixed views, and the patch response. (b) \method{} with one objective removed.}
  \label{fig:ablation}
  \vspace{-8pt}
\end{figure}

\vspace{-4pt}
\section{Conclusion}
\label{sec:conclusion}

In this paper, we study patch-level supervision on aligned pairs for AI-generated image detection. Recent methods mix the two images of a pair and label each patch of the mixed view by its source. We find that patches of different sources in a mixed view interact through self-attention, so the feature of each patch also carries information passed from patches of the other source. A per-patch label ignores this contextual shift and is therefore an imprecise target. Relative Patch Response Learning (\method{}) instead compares the same patch across two mixed views of an aligned pair and learns from the change of its score. Source-unchanged patches respond to the shift alone, so \method{} measures the source-changed patches against them, keeps each group of them moving as a whole, and ranks the two views by generated area. \method{} surpasses the best prior methods by \xstdgain\% and \xwildgain\% in average balanced accuracy on eight standard and three in-the-wild benchmarks, respectively.

\bibliography{references}

@inproceedings{chen2025dda,
  title     = {Dual Data Alignment Makes {AI}-Generated Image Detector Easier Generalizable},
  author    = {Chen, Ruoxin and Xi, Junwei and Yan, Zhiyuan and Zhang, Ke-Yue and Wu, Shuang and Xie, Jingyi and Chen, Xu and Xu, Lei and Guan, Isabel and Yao, Taiping and Ding, Shouhong},
  booktitle = {Advances in Neural Information Processing Systems (NeurIPS)},
  year      = {2025}
}

@inproceedings{ppl2025,
  title     = {All Patches Matter, More Patches Better: Enhance {AI}-Generated Image Detection via Panoptic Patch Learning},
  author    = {Yang, Zheng and Chen, Ruoxin and Yan, Zhiyuan and Zhang, Ke-Yue and Fu, Xinghe and Wu, Shuang and Shu, Xiujun and Yao, Taiping and Ding, Shouhong and Qin, Zequn and Li, Xi},
  booktitle = {International Conference on Learning Representations (ICLR)},
  year      = {2026}
}

@inproceedings{li2025safe,
  title     = {Improving Synthetic Image Detection Towards Generalization: An Image Transformation Perspective},
  author    = {Li, Ouxiang and Cai, Jiayin and Hao, Yanbin and Jiang, Xiaolong and Hu, Yao and Feng, Fuli},
  booktitle = {Proceedings of the ACM SIGKDD Conference on Knowledge Discovery and Data Mining (KDD)},
  year      = {2025}
}

@inproceedings{chen2024drct,
  title     = {{DRCT}: Diffusion Reconstruction Contrastive Training towards Universal Detection of Diffusion Generated Images},
  author    = {Chen, Baoying and Zeng, Jishen and Yang, Jianquan and Yang, Rui},
  booktitle = {International Conference on Machine Learning (ICML)},
  year      = {2024}
}

@inproceedings{rajan2025aligned,
  title     = {Aligned Datasets Improve Detection of Latent Diffusion-Generated Images},
  author    = {Rajan, Anirudh Sundara and Ojha, Utkarsh and Schloesser, Jedidiah and Lee, Yong Jae},
  booktitle = {International Conference on Learning Representations (ICLR)},
  year      = {2025}
}

@inproceedings{guillaro2025bfree,
  title     = {A Bias-Free Training Paradigm for More General {AI}-generated Image Detection},
  author    = {Guillaro, Fabrizio and Zingarini, Giada and Usman, Ben and Sud, Avneesh and Cozzolino, Davide and Verdoliva, Luisa},
  booktitle = {IEEE/CVF Conference on Computer Vision and Pattern Recognition (CVPR)},
  year      = {2025}
}

@inproceedings{ricker2024aeroblade,
  title     = {{AEROBLADE}: Training-Free Detection of Latent Diffusion Images Using Autoencoder Reconstruction Error},
  author    = {Ricker, Jonas and Lukovnikov, Denis and Fischer, Asja},
  booktitle = {IEEE/CVF Conference on Computer Vision and Pattern Recognition (CVPR)},
  year      = {2024}
}

@inproceedings{wang2023dire,
  title     = {{DIRE} for Diffusion-Generated Image Detection},
  author    = {Wang, Zhendong and Bao, Jianmin and Zhou, Wengang and Wang, Weilun and Hu, Hezhen and Chen, Hong and Li, Houqiang},
  booktitle = {IEEE/CVF International Conference on Computer Vision (ICCV)},
  year      = {2023}
}

@inproceedings{wang2020cnnspot,
  title     = {{CNN}-generated Images Are Surprisingly Easy to Spot... for Now},
  author    = {Wang, Sheng-Yu and Wang, Oliver and Zhang, Richard and Owens, Andrew and Efros, Alexei A.},
  booktitle = {IEEE/CVF Conference on Computer Vision and Pattern Recognition (CVPR)},
  year      = {2020}
}

@inproceedings{ojha2023univfd,
  title     = {Towards Universal Fake Image Detectors that Generalize Across Generative Models},
  author    = {Ojha, Utkarsh and Li, Yuheng and Lee, Yong Jae},
  booktitle = {IEEE/CVF Conference on Computer Vision and Pattern Recognition (CVPR)},
  year      = {2023}
}

@inproceedings{tan2024npr,
  title     = {Rethinking the Up-Sampling Operations in {CNN}-based Generative Network for Generalizable Deepfake Detection},
  author    = {Tan, Chuangchuang and Zhao, Yao and Wei, Shikui and Gu, Guanghua and Liu, Ping and Wei, Yunchao},
  booktitle = {IEEE/CVF Conference on Computer Vision and Pattern Recognition (CVPR)},
  year      = {2024}
}

@inproceedings{liu2024fatformer,
  title     = {Forgery-aware Adaptive Transformer for Generalizable Synthetic Image Detection},
  author    = {Liu, Huan and Tan, Zichang and Tan, Chuangchuang and Wei, Yunchao and Wang, Jingdong and Zhao, Yao},
  booktitle = {IEEE/CVF Conference on Computer Vision and Pattern Recognition (CVPR)},
  year      = {2024}
}

@inproceedings{cozzolino2024synthwildx,
  title     = {Raising the Bar of {AI}-generated Image Detection with {CLIP}},
  author    = {Cozzolino, Davide and Poggi, Giovanni and Corvi, Riccardo and Nie{\ss}ner, Matthias and Verdoliva, Luisa},
  booktitle = {IEEE/CVF Conference on Computer Vision and Pattern Recognition Workshops (CVPRW)},
  year      = {2024}
}

@inproceedings{chai2020makes,
  title     = {What Makes Fake Images Detectable? {U}nderstanding Properties that Generalize},
  author    = {Chai, Lucy and Bau, David and Lim, Ser-Nam and Isola, Phillip},
  booktitle = {European Conference on Computer Vision (ECCV)},
  year      = {2020}
}

@inproceedings{corvi2023detection,
  title     = {On the Detection of Synthetic Images Generated by Diffusion Models},
  author    = {Corvi, Riccardo and Cozzolino, Davide and Zingarini, Giada and Poggi, Giovanni and Nagano, Koki and Verdoliva, Luisa},
  booktitle = {IEEE International Conference on Acoustics, Speech and Signal Processing (ICASSP)},
  year      = {2023}
}

@inproceedings{frank2020frequency,
  title     = {Leveraging Frequency Analysis for Deep Fake Image Recognition},
  author    = {Frank, Joel and Eisenhofer, Thorsten and Sch{\"o}nherr, Lea and Fischer, Asja and Kolossa, Dorothea and Holz, Thorsten},
  booktitle = {International Conference on Machine Learning (ICML)},
  year      = {2020}
}

@inproceedings{zhu2023genimage,
  title     = {{GenImage}: A Million-Scale Benchmark for Detecting {AI}-Generated Image},
  author    = {Zhu, Mingjian and Chen, Hanting and Yan, Qiangyu and Huang, Xudong and Lin, Guanyu and Li, Wei and Tu, Zhijun and Hu, Hailin and Hu, Jie and Wang, Yunhe},
  booktitle = {Advances in Neural Information Processing Systems (NeurIPS), Datasets and Benchmarks Track},
  year      = {2023}
}

@inproceedings{yan2025aide,
  title     = {A Sanity Check for {AI}-generated Image Detection},
  author    = {Yan, Shilin and Li, Ouxiang and Cai, Jiayin and Hao, Yanbin and Jiang, Xiaolong and Hu, Yao and Xie, Weidi},
  booktitle = {International Conference on Learning Representations (ICLR)},
  year      = {2025}
}

@article{bammey2023synthbuster,
  title   = {Synthbuster: Towards Detection of Diffusion Model Generated Images},
  author  = {Bammey, Quentin},
  journal = {IEEE Open Journal of Signal Processing},
  volume  = {5},
  pages   = {1--9},
  year    = {2024}
}

@article{cavia2024wildrf,
  title   = {Real-Time Deepfake Detection in the Real-World},
  author  = {Cavia, Bar and Horwitz, Eliahu and Reiss, Tal and Hoshen, Yedid},
  journal = {arXiv preprint arXiv:2406.09398},
  year    = {2024}
}

@article{simeoni2025dinov3,
  title   = {{DINOv3}},
  author  = {Sim{\'e}oni, Oriane and Vo, Huy V. and Seitzer, Maximilian and Baldassarre, Federico and Oquab, Maxime and Jose, Cijo and Khalidov, Vasil and Szafraniec, Marc and Yi, Seungeun and Ramamonjisoa, Micha{\"e}l and others},
  journal = {arXiv preprint arXiv:2508.10104},
  year    = {2025}
}

@inproceedings{hu2022lora,
  title     = {{LoRA}: Low-Rank Adaptation of Large Language Models},
  author    = {Hu, Edward J. and Shen, Yelong and Wallis, Phillip and Allen-Zhu, Zeyuan and Li, Yuanzhi and Wang, Shean and Wang, Lu and Chen, Weizhu},
  booktitle = {International Conference on Learning Representations (ICLR)},
  year      = {2022}
}

@inproceedings{jia2023autosplice,
  title     = {{AutoSplice}: A Text-prompt Manipulated Image Dataset for Media Forensics},
  author    = {Jia, Shan and Huang, Mingzhen and Zhou, Zhou and Ju, Yan and Cai, Jialing and Lyu, Siwei},
  booktitle = {IEEE/CVF Conference on Computer Vision and Pattern Recognition Workshops (CVPRW)},
  year      = {2023},
}

@inproceedings{qin2026gapl,
  title     = {Scaling Up {AI}-Generated Image Detection with Generator-Aware Prototypes},
  author    = {Qin, Ziheng and Ji, Yuheng and Tao, Renshuai and Tian, Yuxuan and Liu, Yuyang and Wang, Yipu and Zheng, Xiaolong},
  booktitle = {IEEE/CVF Conference on Computer Vision and Pattern Recognition (CVPR)},
  year      = {2026}
}

@inproceedings{tan2025c2pclip,
  title     = {{C2P-CLIP}: Injecting Category Common Prompt in {CLIP} to Enhance Generalization in Deepfake Detection},
  author    = {Tan, Chuangchuang and Tao, Renshuai and Liu, Huan and Gu, Guanghua and Wu, Baoyuan and Zhao, Yao and Wei, Yunchao},
  booktitle = {AAAI Conference on Artificial Intelligence (AAAI)},
  year      = {2025},
}

@inproceedings{yan2025effort,
  title     = {Orthogonal Subspace Decomposition for Generalizable {AI}-Generated Image Detection},
  author    = {Yan, Zhiyuan and Wang, Jiangming and Jin, Peng and Zhang, Ke-Yue and Liu, Chengchun and Chen, Shen and Yao, Taiping and Ding, Shouhong and Wu, Baoyuan and Yuan, Li},
  booktitle = {International Conference on Machine Learning (ICML)},
  year      = {2025},
}

@article{zhong2023patchcraft,
  title   = {{PatchCraft}: Exploring Texture Patch for Efficient {AI}-generated Image Detection},
  author  = {Zhong, Nan and Xu, Yiran and Li, Sheng and Qian, Zhenxing and Zhang, Xinpeng},
  journal = {arXiv preprint arXiv:2311.12397},
  year    = {2023},
}

@inproceedings{lin2014coco,
  title     = {Microsoft {COCO}: Common Objects in Context},
  author    = {Lin, Tsung-Yi and Maire, Michael and Belongie, Serge and Hays, James and Perona, Pietro and Ramanan, Deva and Doll{\'a}r, Piotr and Zitnick, C. Lawrence},
  booktitle = {European Conference on Computer Vision (ECCV)},
  year      = {2014}
}

@inproceedings{rombach2022ldm,
  title     = {High-Resolution Image Synthesis with Latent Diffusion Models},
  author    = {Rombach, Robin and Blattmann, Andreas and Lorenz, Dominik and Esser, Patrick and Ommer, Bj{\"o}rn},
  booktitle = {IEEE/CVF Conference on Computer Vision and Pattern Recognition (CVPR)},
  year      = {2022}
}

@inproceedings{kingma2015adam,
  title     = {Adam: A Method for Stochastic Optimization},
  author    = {Kingma, Diederik P. and Ba, Jimmy},
  booktitle = {International Conference on Learning Representations (ICLR)},
  year      = {2015}
}

@article{dailybench2026,
  title   = {{DailyBench}: A Unified Benchmark for {AI}-Generated and Manipulated Images from Modern Generative Models},
  author  = {Jiang, Xin and Tang, Hao and Gao, Junyao and Yang, Zijie and Cao, Meiqi and Shen, Fei and Zhang, Dongming and Li, Zechao and Zhang, Yongdong},
  journal = {arXiv preprint arXiv:2607.24016},
  year    = {2026}
}

@inproceedings{karras2019stylegan,
  title     = {A Style-Based Generator Architecture for Generative Adversarial Networks},
  author    = {Karras, Tero and Laine, Samuli and Aila, Timo},
  booktitle = {IEEE/CVF Conference on Computer Vision and Pattern Recognition (CVPR)},
  year      = {2019},
}

@inproceedings{dhariwal2021diffusion,
  title     = {Diffusion Models Beat {GANs} on Image Synthesis},
  author    = {Dhariwal, Prafulla and Nichol, Alexander},
  booktitle = {Advances in Neural Information Processing Systems (NeurIPS)},
  year      = {2021},
}

@inproceedings{koutlis2024rine,
  title     = {Leveraging Representations from Intermediate Encoder-Blocks for Synthetic Image Detection},
  author    = {Koutlis, Christos and Papadopoulos, Symeon},
  booktitle = {European Conference on Computer Vision (ECCV)},
  year      = {2024},
}

@inproceedings{tan2024freqnet,
  title     = {Frequency-Aware Deepfake Detection: Improving Generalizability through Frequency Space Domain Learning},
  author    = {Tan, Chuangchuang and Zhao, Yao and Wei, Shikui and Gu, Guanghua and Liu, Ping and Wei, Yunchao},
  booktitle = {AAAI Conference on Artificial Intelligence (AAAI)},
  volume    = {38},
  pages     = {5052--5060},
  year      = {2024}
}

@inproceedings{karageorgiou2025spai,
  title     = {Any-Resolution {AI}-Generated Image Detection by Spectral Learning},
  author    = {Karageorgiou, Dimitrios and Papadopoulos, Symeon and Kompatsiaris, Ioannis and Gavves, Efstratios},
  booktitle = {IEEE/CVF Conference on Computer Vision and Pattern Recognition (CVPR)},
  pages     = {18706--18717},
  year      = {2025}
}

@inproceedings{luo2024lare,
  title     = {{LaRE}\textsuperscript{2}: Latent Reconstruction Error Based Method for Diffusion-Generated Image Detection},
  author    = {Luo, Yunpeng and Du, Junlong and Yan, Ke and Ding, Shouhong},
  booktitle = {IEEE/CVF Conference on Computer Vision and Pattern Recognition (CVPR)},
  pages     = {17006--17015},
  year      = {2024}
}

@inproceedings{chu2025fire,
  title     = {{FIRE}: Robust Detection of Diffusion-Generated Images via Frequency-Guided Reconstruction Error},
  author    = {Chu, Beilin and Xu, Xuan and Wang, Xin and Zhang, Yufei and You, Weike and Zhou, Linna},
  booktitle = {IEEE/CVF Conference on Computer Vision and Pattern Recognition (CVPR)},
  year      = {2025}
}

@inproceedings{park2025community,
  title     = {Community Forensics: Using Thousands of Generators to Train Fake Image Detectors},
  author    = {Park, Jeongsoo and Owens, Andrew},
  booktitle = {IEEE/CVF Conference on Computer Vision and Pattern Recognition (CVPR)},
  pages     = {8245--8257},
  year      = {2025}
}

@inproceedings{yang2025d3,
  title     = {{D}\textsuperscript{3}: Scaling Up Deepfake Detection by Learning from Discrepancy},
  author    = {Yang, Yongqi and Qian, Zhihao and Zhu, Ye and Russakovsky, Olga and Wu, Yu},
  booktitle = {IEEE/CVF Conference on Computer Vision and Pattern Recognition (CVPR)},
  pages     = {23850--23859},
  year      = {2025}
}

@inproceedings{yu2024semgir,
  title     = {{SemGIR}: Semantic-Guided Image Regeneration Based Method for {AI}-generated Image Detection and Attribution},
  author    = {Yu, Xiao and Chen, Kejiang and Zeng, Kai and Fang, Han and Yang, Zijin and Shang, Xiuwei and Qi, Yuang and Zhang, Weiming and Yu, Nenghai},
  booktitle = {ACM International Conference on Multimedia (ACM MM)},
  pages     = {8480--8488},
  year      = {2024}
}

@inproceedings{grommelt2024jpeg,
  title     = {Fake or {JPEG}? {R}evealing Common Biases in Generated Image Detection Datasets},
  author    = {Grommelt, Patrick and Weiss, Louis and Pfreundt, Franz-Josef and Keuper, Janis},
  booktitle = {European Conference on Computer Vision Workshops (ECCVW)},
  pages     = {80--95},
  year      = {2024}
}

@inproceedings{esser2024sd3,
  title     = {Scaling Rectified Flow Transformers for High-Resolution Image Synthesis},
  author    = {Esser, Patrick and Kulal, Sumith and Blattmann, Andreas and Entezari, Rahim and M{\"u}ller, Jonas and Saini, Harry and Levi, Yam and Lorenz, Dominik and Sauer, Axel and Boesel, Frederic and Podell, Dustin and Dockhorn, Tim and English, Zion and Rombach, Robin},
  booktitle = {International Conference on Machine Learning (ICML)},
  year      = {2024},
}

@article{labs2025kontext,
  title   = {{FLUX.1 Kontext}: Flow Matching for In-Context Image Generation and Editing in Latent Space},
  author  = {{Black Forest Labs} and Batifol, Stephen and Blattmann, Andreas and Boesel, Frederic and others},
  journal = {arXiv preprint arXiv:2506.15742},
  year    = {2025}
}

@article{wu2025qwenimage,
  title   = {{Qwen-Image} Technical Report},
  author  = {Wu, Chenfei and Li, Jiahao and Zhou, Jingren and others},
  journal = {arXiv preprint arXiv:2508.02324},
  year    = {2025}
}

@misc{openai2025imagegen,
  title        = {Addendum to {GPT-4o} System Card: Native Image Generation},
  author       = {{OpenAI}},
  year         = {2025},
  howpublished = {\url{https://openai.com/index/gpt-4o-image-generation-system-card-addendum/}}
}

@inproceedings{lu2023seeing,
  title     = {Seeing Is Not Always Believing: Benchmarking Human and Model Perception of {AI}-Generated Images},
  author    = {Lu, Zeyu and Huang, Di and Bai, Lei and Qu, Jingjing and Wu, Chengyue and Liu, Xihui and Ouyang, Wanli},
  booktitle = {Advances in Neural Information Processing Systems (NeurIPS), Datasets and Benchmarks Track},
  year      = {2023}
}

@article{nightingale2022faces,
  title   = {{AI}-Synthesized Faces Are Indistinguishable from Real Faces and More Trustworthy},
  author  = {Nightingale, Sophie J. and Farid, Hany},
  journal = {Proceedings of the National Academy of Sciences},
  volume  = {119},
  number  = {8},
  pages   = {e2120481119},
  year    = {2022},
}

@article{feuerriegel2023disinformation,
  title   = {Research Can Help to Tackle {AI}-Generated Disinformation},
  author  = {Feuerriegel, Stefan and DiResta, Ren{\'e}e and Goldstein, Josh A. and Kumar, Srijan and Lorenz-Spreen, Philipp and Tomz, Michael and Pr{\"o}llochs, Nicolas},
  journal = {Nature Human Behaviour},
  volume  = {7},
  pages   = {1818--1821},
  year    = {2023}
}

@inproceedings{chen2025syntheticpolitics,
  title     = {Synthetic Politics: Prevalence, Spreaders, and Emotional Reception of {AI}-Generated Political Images on {X}},
  author    = {Chen, Zhiyi and Ye, Jinyi and Tsai, Beverlyn and Ferrara, Emilio and Luceri, Luca},
  booktitle = {ACM Conference on Hypertext and Social Media (HT)},
  year      = {2025}
}

@article{yang2024fakeprofiles,
  title   = {Characteristics and Prevalence of Fake Social Media Profiles with {AI}-Generated Faces},
  author  = {Yang, Kai-Cheng and Singh, Danishjeet and Menczer, Filippo},
  journal = {Journal of Online Trust and Safety},
  volume  = {2},
  number  = {4},
  year    = {2024}
}

@inproceedings{dhakal2026simlbr,
  title     = {{SimLBR}: Learning to Detect Fake Images by Learning to Detect Real Images},
  author    = {Dhakal, Aayush and Khanal, Subash and Sastry, Srikumar and Arndt, Jacob and Dias, Philipe Ambrozio and Lunga, Dalton and Jacobs, Nathan},
  booktitle = {IEEE/CVF Conference on Computer Vision and Pattern Recognition (CVPR)},
  year      = {2026}
}

@inproceedings{li2026cptformer,
  title     = {Detecting Compressed {AI}-Generated Images via Phase Spectrum Robustness},
  author    = {Li, Kai and Ren, Wenqi and Wang, Wei and Cao, Xiaochun},
  booktitle = {IEEE/CVF Conference on Computer Vision and Pattern Recognition (CVPR)},
  year      = {2026},
}

@inproceedings{fu2025pid,
  title     = {{PiD}: Generalized {AI}-Generated Images Detection with Pixelwise Decomposition Residuals},
  author    = {Fu, Xinghe and Yan, Zhiyuan and Yang, Zheng and Yao, Taiping and Zhao, Yandan and Ding, Shouhong and Li, Xi},
  booktitle = {International Conference on Machine Learning (ICML)},
  year      = {2025}
}

@inproceedings{qi2026did,
  title     = {A Difference-in-Difference Approach to Detecting {AI}-Generated Images},
  author    = {Qi, Xinyi and Ye, Kai and Shi, Chengchun and Yang, Ying and Zhu, Jin and Zhou, Hongyi},
  booktitle = {IEEE/CVF Conference on Computer Vision and Pattern Recognition (CVPR)},
  year      = {2026}
}

@inproceedings{wu2026rawrgb,
  title     = {Zero-shot Detection of {AI}-Generated Image via {RAW}-{RGB} Alignment},
  author    = {Wu, Haiwei and Li, Fengpeng and Tu, Zhilin and Li, Yuanman and Li, Xiong and Zhou, Jiantao},
  booktitle = {IEEE/CVF Conference on Computer Vision and Pattern Recognition (CVPR)},
  year      = {2026}
}

@inproceedings{li2026easier,
  title     = {Easier Painting Than Thinking: Can Text-to-Image Models Set the Stage, but Not Direct the Play?},
  author    = {Li, Ouxiang and Wang, Yuan and Hu, Xinting and Huang, Huijuan and Chen, Rui and Ou, Jiarong and Tao, Xin and Wan, Pengfei and Qi, Xiaojuan and Feng, Fuli},
  booktitle = {International Conference on Learning Representations (ICLR)},
  pages     = {86729--86758},
  year      = {2026}
}
\bibliographystyle{prl_preprint}

\clearpage
\appendix
\section{Implementation Details}
\label{app:impl}

\paragraph{Architecture.}
The frozen DINOv3 ViT-L/16 backbone is adapted with LoRA of rank $8$ and scaling $\alpha=1$, applied in every transformer block to the query, key, value and output projections of self-attention and to the two projections of the MLP. The patch scoring head is a single linear layer on the final-layer patch tokens. The detection head is a gating MLP that assigns a weight to every token, aggregates the tokens with the normalized weights and maps the result to the image logit.

\paragraph{Data augmentation.}
At both training and inference, we center-crop every image to $336\times336$ pixels, and we resize an image to the input size instead when both its width and height exceed $1{,}024$ pixels, so that the input covers the whole image. We apply the common augmentations in this field, such as random JPEG compression, Gaussian blur and colour jitter. Each augmentation is drawn once per pair and applied identically to both members, so that the alignment is preserved.

\paragraph{View sampling.}
For each training pair, with probability $0.5$ we use the real image with two mixed views, and otherwise the real image with the synthetic image. Each mask replaces one axis-aligned rectangle of patches on the $21\times21$ patch grid of the encoder, so $P=441$. The two masks of a pair share a base image, the real or the synthetic one with equal probability, and the rectangle takes its patches from the other image. For each mask, we draw the synthetic fraction uniformly from $[0.1, 0.9]$ and the aspect ratio of the rectangle log-uniformly from $[0.5, 2]$, resample until the two fractions differ, and place the rectangle uniformly at random among the positions where it fits.

\section{Benchmarks}
\label{app:datasets}

\paragraph{GenImage}~\citep{zhu2023genimage} pairs ImageNet photographs with images from eight generators: ADM, BigGAN, Glide, Midjourney, Stable Diffusion v1.4 and v1.5, VQDM and Wukong.

\paragraph{AIGCDetectBenchmark}~\citep{zhong2023patchcraft} collects seventeen subsets from GANs (ProGAN, StyleGAN, StyleGAN2, BigGAN, CycleGAN, StarGAN, GauGAN and the WFIR face set) and diffusion models (ADM, Glide, Midjourney, Stable Diffusion v1.4, v1.5 and XL, VQDM, Wukong and DALL-E~2), with real images drawn from LSUN, ImageNet, MSCOCO, CelebA and FFHQ.

\paragraph{DRCT-2M}~\citep{chen2024drct} generates images from MSCOCO with sixteen diffusion models. Ten text-to-image models (LDM, SD1.4, SD1.5, SD2.1, SDXL, SDXL-Refiner, SD-Turbo, SDXL-Turbo, LCM-SD1.5 and LCM-SDXL) are prompted by the MSCOCO captions, three ControlNet models are conditioned on the captions and Canny edges, and three inpainting models reconstruct the real images with an empty prompt and an all-zero mask (SDv1-DR, SDv2-DR and SDXL-DR). Every subset is paired with the shared real set.

\paragraph{DDA-COCO and EvalGEN}~\citep{chen2025dda} are released with DDA. DDA-COCO reconstructs the MSCOCO validation images with six VAEs, including those of SD1.x, SD2.1, SDXL, SD3.5 and Flux, with frequency-level alignment to the real images. EvalGEN holds $2{,}765$ images from five recent text-to-image generators (Flux, GoT, Infinity, OmniGen and NOVA) driven by GenEval prompts; it has no real counterpart, so we report accuracy on its synthetic images.

\paragraph{Synthbuster}~\citep{bammey2023synthbuster} provides $1{,}000$ images from each of nine generators (DALL-E~2, DALL-E~3, Firefly, Glide, Midjourney~v5, Stable Diffusion 1.3, 1.4, 2 and XL).

\paragraph{ForenSynths}~\citep{wang2020cnnspot} is the test set of CNNSpot, with thirteen subsets from CNN-based generators: ProGAN, StyleGAN, StyleGAN2, BigGAN, CycleGAN, StarGAN, GauGAN, CRN, IMLE, SITD, SAN, DeepFake and WFIR.

\paragraph{UnivFD}~\citep{ojha2023univfd} extends the eleven ForenSynths subsets with eight diffusion and auto-regressive subsets: Guided, three LDM variants, three Glide variants and DALL-E, nineteen subsets in total.

\paragraph{Chameleon}~\citep{yan2025aide} collects about $26{,}000$ real and generated images from the web. The generated images come from AI-art communities and passed a human perception test, every image is at least $448\times448$ pixels, and the generators are unknown.

\paragraph{SynthWildX}~\citep{cozzolino2024synthwildx} holds $500$ real and $1{,}500$ generated images collected from X, attributed by the tags and annotations on the platform to DALL-E~3, Midjourney and Firefly, which form its three subsets.

\paragraph{WildRF}~\citep{cavia2024wildrf} collects real and generated images from Reddit, Facebook and X, one subset per platform, with unknown generators and varying resolutions, formats, compressions and editing transformations.

\paragraph{AutoSplice}~\citep{jia2023autosplice} replaces one object region of a real photograph by a DALL-E~2 inpainting guided by a text prompt, and releases the forged images at JPEG qualities $100$, $90$ and $75$.

\paragraph{DailyBench}~\citep{dailybench2026} collects edited versions of real images from LAION-Aesthetics; its ManipulationBench subset, which we use, contains edits by four open-source editors (FLUX-Fill with object masks or random rectangular masks, FLUX.2-klein-9B, Qwen-Image-Edit-2511 and Step1X-Edit-v1p2) and two proprietary ones (Nano Banana~2 and GPT-Image~2), mostly object removals and replacements.

\section{Compared Methods}
\label{app:methods}

\paragraph{UnivFD}~\citep{ojha2023univfd} keeps the CLIP ViT-L/14 image encoder frozen and trains only a linear classifier on its features, arguing that a feature space not tuned to the training generator generalizes better.

\paragraph{FatFormer}~\citep{liu2024fatformer} adds forgery-aware adapters with a frequency branch to the CLIP image encoder and aligns the adapted image features with text prompts.

\paragraph{NPR}~\citep{tan2024npr} models the relations between neighbouring pixels that up-sampling layers of generators leave behind and feeds them to a ResNet classifier.

\paragraph{DRCT}~\citep{chen2024drct} reconstructs both real and generated images with a diffusion model to obtain hard samples of the same content, labels every reconstruction as generated, and trains a detector with a classification and a contrastive loss.

\paragraph{C2P-CLIP}~\citep{tan2025c2pclip} injects category-common prompts into the CLIP image encoder through the text encoder, so that the image features carry the concepts of real and generated images.

\paragraph{AIDE}~\citep{yan2025aide} combines the highest- and lowest-frequency patches of an image, selected by DCT scoring and processed by a CNN, with the semantic features of CLIP; the paper also introduces the Chameleon benchmark.

\paragraph{AlignedForensics}~\citep{rajan2025aligned} pairs every real image with its VAE reconstruction so that the two differ only in the traces of generation, and trains a detector on these aligned images.

\paragraph{Effort}~\citep{yan2025effort} decomposes the weights of the CLIP encoder into orthogonal subspaces by singular value decomposition, freezes the principal subspace to preserve the pretrained knowledge and trains the residual subspace for detection.

\paragraph{SAFE}~\citep{li2025safe} feeds the high-frequency components of a discrete wavelet transform to a lightweight ResNet and pairs it with image transformations that preserve the traces of generation: cropping instead of resizing, colour jitter, random rotation and random masking.

\paragraph{DDA}~\citep{chen2025dda} aligns generated images with real ones in both the pixel and the frequency domain: VAE reconstruction, high-frequency debias from the real image and pixel mixup.

\paragraph{GAPL}~\citep{qin2026gapl} builds generator-aware prototypes from the principal components of the embeddings of a few generators and fine-tunes the CLIP ViT-L/14 encoder with LoRA so that image embeddings are mapped onto these prototypes. It is trained on the Community Forensics set~\citep{park2025community} of several thousand generators.

\section{Detailed Benchmark Results}
\label{app:results}
\makeatletter\setlength{\@fptop}{0pt}\setlength{\@fpsep}{12pt}\makeatother

Tables~\ref{tab:genimage}, \ref{tab:aigcdb}, \ref{tab:synthbuster}, \ref{tab:drct}, \ref{tab:ddacoco_evalgen} and \ref{tab:forensynths} report the per-subset results on GenImage, AIGCDetect, Synthbuster, DRCT-2M, DDA-COCO and EvalGEN, and ForenSynths, which Table~\ref{tab:main} and Section~\ref{sec:main} summarize. \method{} is above $98.8\%$ on every DRCT-2M subset and above $99\%$ on every EvalGEN generator; on DDA-COCO its two lowest subsets are the SD3.5 and Flux reconstructions. On ForenSynths, its margin over the alignment-based methods opens on every GAN generator, while the DeepFake, SITD and SAN subsets remain its hardest.

Table~\ref{tab:dailybench} reports balanced accuracy on the ManipulationBench subset of DailyBench (Section~\ref{sec:local}), with the rows of the other detectors, including the detector FPD proposed with the benchmark, as published by DailyBench. Most of the detectors it evaluates stay within a few points of chance on most editors. \method{} has the best average, $68.32\%$ against $66.88\%$ for DDA and $63.41\%$ for FPD, and the best result on the three FLUX-based editors, where every other detector except DDA is at chance. DDA is stronger on Qwen-Image-Edit and Step1X-Edit, and SAFE on GPT-Image~2.

\begin{table}[!t]
  \caption{Comparison of balanced accuracy (\%) on GenImage.}
  \label{tab:genimage}
  \centering
  \renewcommand{\arraystretch}{1.25}
  \scriptsize
  \setlength{\tabcolsep}{3pt}
\resizebox{\ifdim\width>\textwidth\textwidth\else\width\fi}{!}{\scalebox{\tabscale}{%
  \begin{tabular}{lccccccccc}
    \toprule\addlinespace[4pt]
    Method & ADM & BigGAN & Glide & Midjourney & SDv1.4 & SDv1.5 & VQDM & Wukong & \textbf{Avg.} \\
    \midrule
    UnivFD\,\raisebox{-0.25ex}{\textcolor{venuecolor}{\tiny CVPR'23}} & 62.51 & 84.37 & 61.28 & 55.06 & 55.57 & 55.67 & 76.88 & 61.10 & 64.05 \\
    FatFormer\,\raisebox{-0.25ex}{\textcolor{venuecolor}{\tiny CVPR'24}} & 59.82 & 81.77 & 60.28 & 51.99 & 52.81 & 52.76 & 69.97 & 58.86 & 61.03 \\
    NPR\,\raisebox{-0.25ex}{\textcolor{venuecolor}{\tiny CVPR'24}} & 46.73 & 44.25 & 46.06 & 43.86 & 46.73 & 47.06 & 50.00 & 49.22 & 46.74 \\
    DRCT\,\raisebox{-0.25ex}{\textcolor{venuecolor}{\tiny ICML'24}} & 60.74 & 58.38 & 67.86 & 88.68 & \underline{99.37} & 99.21 & 70.03 & 99.10 & 80.42 \\
    C2P-CLIP\,\raisebox{-0.25ex}{\textcolor{venuecolor}{\tiny AAAI'25}} & 53.62 & \underline{92.67} & 65.91 & 50.60 & 55.02 & 55.31 & 63.58 & 60.00 & 62.09 \\
    AIDE\,\raisebox{-0.25ex}{\textcolor{venuecolor}{\tiny ICLR'25}} & 50.22 & 50.42 & 52.26 & 57.46 & 75.96 & 76.19 & 50.78 & 69.47 & 60.35 \\
    AlignedForensics\,\raisebox{-0.25ex}{\textcolor{venuecolor}{\tiny ICLR'25}} & 50.49 & 52.73 & 52.80 & \underline{96.33} & \textbf{99.82} & \textbf{99.74} & 67.83 & \textbf{99.79} & 77.44 \\
    Effort\,\raisebox{-0.25ex}{\textcolor{venuecolor}{\tiny ICML'25}} & 69.75 & 51.60 & 83.43 & 76.37 & 96.03 & 95.82 & 73.24 & 90.17 & 79.55 \\
    SAFE\,\raisebox{-0.25ex}{\textcolor{venuecolor}{\tiny KDD'25}} & 50.41 & 51.84 & 53.73 & 49.89 & 50.55 & 50.54 & 51.01 & 51.15 & 51.14 \\
    DDA\,\raisebox{-0.25ex}{\textcolor{venuecolor}{\tiny NeurIPS'25}} & 89.55 & 86.52 & 89.58 & 95.60 & 98.72 & 98.55 & 76.58 & 98.78 & 91.74 \\
    GAPL\,\raisebox{-0.25ex}{\textcolor{venuecolor}{\tiny CVPR'26}} & \textbf{94.46} & \textbf{97.17} & \textbf{97.52} & 84.41 & 97.81 & 97.59 & \underline{97.69} & 97.81 & \underline{95.56} \\
    \midrule
    \rowcolor{oursrow} \textbf{\method{} (Ours)} & \underline{94.40} & 91.55 & \underline{95.02} & \textbf{96.73} & 99.33 & \underline{99.25} & \textbf{99.05} & \underline{99.36} & \textbf{96.83} \\
    \addlinespace[4pt]\bottomrule
  \end{tabular}}}
\end{table}

\begin{table}[!t]
  \caption{Comparison of balanced accuracy (\%) on AIGCDetectBenchmark.}
  \label{tab:aigcdb}
  \centering
  \renewcommand{\arraystretch}{1.25}
  \scriptsize
  \setlength{\tabcolsep}{3pt}
\resizebox{\ifdim\width>\textwidth\textwidth\else\width\fi}{!}{\scalebox{\tabscale}{%
  \begin{tabular}{lcccccccccccccccccc}
    \toprule\addlinespace[4pt]
    Method & ADM & BigGAN & CycleGAN & DALLE2 & GauGAN & Glide & Midjourney & ProGAN & SDXL & SDv1.4 & SDv1.5 & StarGAN & StyleGAN & StyleGAN2 & VQDM & WFIR & Wukong & \textbf{Avg.} \\
    \midrule
    UnivFD\,\raisebox{-0.25ex}{\textcolor{venuecolor}{\tiny CVPR'23}} & 62.51 & 87.50 & 96.92 & 49.95 & 98.79 & 61.28 & 55.06 & \underline{99.38} & 58.15 & 55.57 & 55.67 & 95.12 & 79.99 & 69.41 & 76.88 & 69.20 & 61.10 & 72.50 \\
    FatFormer\,\raisebox{-0.25ex}{\textcolor{venuecolor}{\tiny CVPR'24}} & 59.82 & 97.90 & \underline{99.32} & 49.00 & 99.17 & 60.28 & 51.99 & 99.21 & 63.75 & 52.81 & 52.76 & \textbf{99.75} & 87.51 & 81.63 & 69.97 & 87.50 & 58.86 & 74.78 \\
    NPR\,\raisebox{-0.25ex}{\textcolor{venuecolor}{\tiny CVPR'24}} & 46.73 & 43.25 & 59.83 & 46.65 & 35.62 & 46.06 & 43.86 & 50.45 & 47.48 & 46.73 & 47.06 & 49.47 & 50.18 & 50.02 & 50.00 & 49.65 & 49.22 & 47.78 \\
    DRCT\,\raisebox{-0.25ex}{\textcolor{venuecolor}{\tiny ICML'24}} & 60.74 & 52.45 & 48.83 & 68.25 & 49.45 & 67.86 & 88.68 & 53.02 & 80.65 & \underline{99.37} & 99.21 & 54.38 & 54.45 & 53.74 & 70.03 & 50.15 & 99.10 & 67.67 \\
    C2P-CLIP\,\raisebox{-0.25ex}{\textcolor{venuecolor}{\tiny AAAI'25}} & 53.62 & 92.07 & \textbf{99.39} & 50.20 & 98.20 & 65.91 & 50.60 & 97.10 & 51.20 & 55.02 & 55.31 & \underline{99.70} & 76.97 & 58.51 & 63.58 & 65.35 & 60.00 & 70.16 \\
    AIDE\,\raisebox{-0.25ex}{\textcolor{venuecolor}{\tiny ICLR'25}} & 50.22 & 43.68 & 51.73 & 50.95 & 39.73 & 52.26 & 57.46 & 49.31 & 50.55 & 75.96 & 76.19 & 51.43 & 49.57 & 49.20 & 50.78 & 49.20 & 69.47 & 53.98 \\
    AlignedForensics\,\raisebox{-0.25ex}{\textcolor{venuecolor}{\tiny ICLR'25}} & 50.49 & 50.52 & 49.69 & 50.30 & 50.24 & 52.80 & \underline{96.33} & 50.35 & 94.45 & \textbf{99.82} & \textbf{99.74} & 51.50 & 51.46 & 50.31 & 67.83 & 50.00 & \textbf{99.79} & 65.63 \\
    Effort\,\raisebox{-0.25ex}{\textcolor{venuecolor}{\tiny ICML'25}} & 69.75 & 73.60 & 81.49 & 79.15 & 62.52 & 83.43 & 76.37 & 70.08 & 86.88 & 96.03 & 95.82 & 96.02 & 56.42 & 47.67 & 73.24 & 70.35 & 90.17 & 77.00 \\
    SAFE\,\raisebox{-0.25ex}{\textcolor{venuecolor}{\tiny KDD'25}} & 50.41 & 48.95 & 52.58 & 50.05 & 42.60 & 53.73 & 49.89 & 49.96 & 49.93 & 50.55 & 50.54 & 49.95 & 49.99 & 49.99 & 51.01 & 49.10 & 51.15 & 50.02 \\
    DDA\,\raisebox{-0.25ex}{\textcolor{venuecolor}{\tiny NeurIPS'25}} & 89.55 & 90.97 & 72.47 & \underline{94.55} & 92.74 & 89.58 & 95.60 & 92.80 & \underline{99.38} & 98.72 & 98.55 & 72.74 & 87.80 & 90.17 & 76.58 & 52.10 & 98.78 & 87.83 \\
    GAPL\,\raisebox{-0.25ex}{\textcolor{venuecolor}{\tiny CVPR'26}} & \textbf{94.46} & \underline{98.62} & 97.54 & 84.05 & \textbf{99.63} & \textbf{97.52} & 84.41 & \textbf{99.83} & 98.78 & 97.81 & 97.59 & 97.05 & \underline{97.89} & \textbf{98.62} & \underline{97.69} & \textbf{94.10} & 97.81 & \underline{96.08} \\
    \midrule
    \rowcolor{oursrow} \textbf{\method{} (Ours)} & \underline{94.40} & \textbf{99.28} & 97.37 & \textbf{96.55} & \underline{99.55} & \underline{95.02} & \textbf{96.73} & 98.99 & \textbf{99.55} & 99.33 & \underline{99.25} & 99.25 & \textbf{98.21} & \underline{98.19} & \textbf{99.05} & \underline{91.65} & \underline{99.36} & \textbf{97.75} \\
    \addlinespace[4pt]\bottomrule
  \end{tabular}}}
\end{table}

\begin{table}[!t]
  \caption{Comparison of balanced accuracy (\%) on Synthbuster.}
  \label{tab:synthbuster}
  \centering
  \renewcommand{\arraystretch}{1.25}
  \scriptsize
  \setlength{\tabcolsep}{3pt}
\resizebox{\ifdim\width>\textwidth\textwidth\else\width\fi}{!}{\scalebox{\tabscale}{%
  \begin{tabular}{lcccccccccc}
    \toprule\addlinespace[4pt]
    Method & DALL-E~2 & DALL-E~3 & Firefly & GLIDE & Midjourney & SD1.3 & SD1.4 & SD2 & SDXL & \textbf{Avg.} \\
    \midrule
    UnivFD\,\raisebox{-0.25ex}{\textcolor{venuecolor}{\tiny CVPR'23}} & 70.98 & 34.83 & 77.38 & 40.73 & 39.93 & 57.88 & 57.38 & 63.13 & 55.48 & 55.31 \\
    FatFormer\,\raisebox{-0.25ex}{\textcolor{venuecolor}{\tiny CVPR'24}} & 53.53 & 35.33 & 59.83 & 67.28 & 43.88 & 52.43 & 53.18 & 51.03 & 66.08 & 53.62 \\
    NPR\,\raisebox{-0.25ex}{\textcolor{venuecolor}{\tiny CVPR'24}} & 55.76 & 12.36 & 9.46 & 53.01 & 40.81 & 55.76 & 56.41 & 35.86 & 57.11 & 41.83 \\
    DRCT\,\raisebox{-0.25ex}{\textcolor{venuecolor}{\tiny ICML'24}} & 47.54 & 53.39 & 48.59 & 60.19 & 89.29 & 94.09 & 94.09 & 90.79 & 79.39 & 73.04 \\
    C2P-CLIP\,\raisebox{-0.25ex}{\textcolor{venuecolor}{\tiny AAAI'25}} & 54.89 & 43.64 & 81.54 & 81.59 & 52.74 & 79.69 & 79.94 & 65.64 & 72.99 & 68.08 \\
    AIDE\,\raisebox{-0.25ex}{\textcolor{venuecolor}{\tiny ICLR'25}} & 37.83 & 36.63 & 27.73 & 67.93 & 60.48 & 77.03 & 76.68 & 56.18 & 71.38 & 56.87 \\
    AlignedForensics\,\raisebox{-0.25ex}{\textcolor{venuecolor}{\tiny ICLR'25}} & 49.15 & 48.60 & 50.00 & 56.60 & \textbf{98.55} & \textbf{98.60} & \textbf{98.60} & \textbf{98.30} & \underline{97.25} & 77.29 \\
    Effort\,\raisebox{-0.25ex}{\textcolor{venuecolor}{\tiny ICML'25}} & 67.88 & 79.28 & 56.08 & 78.98 & 77.23 & 81.63 & 81.68 & 77.88 & 80.48 & 75.68 \\
    SAFE\,\raisebox{-0.25ex}{\textcolor{venuecolor}{\tiny KDD'25}} & 64.27 & 16.22 & 16.57 & 58.47 & 63.02 & 65.72 & 65.42 & 59.32 & 65.77 & 52.75 \\
    DDA\,\raisebox{-0.25ex}{\textcolor{venuecolor}{\tiny NeurIPS'25}} & 81.79 & \underline{85.44} & 87.39 & 71.94 & 88.94 & 88.34 & 88.14 & 88.79 & 88.94 & 85.52 \\
    GAPL\,\raisebox{-0.25ex}{\textcolor{venuecolor}{\tiny CVPR'26}} & \underline{91.69} & 74.39 & \underline{91.54} & \underline{93.29} & 90.64 & 93.54 & 93.59 & 92.64 & 94.29 & \underline{90.63} \\
    \midrule
    \rowcolor{oursrow} \textbf{\method{} (Ours)} & \textbf{97.65} & \textbf{97.85} & \textbf{93.40} & \textbf{94.85} & \underline{97.85} & \underline{97.85} & \underline{97.85} & \underline{97.85} & \textbf{97.85} & \textbf{97.00} \\
    \addlinespace[4pt]\bottomrule
  \end{tabular}}}
\end{table}

\begin{table}[!t]
  \caption{Comparison of balanced accuracy (\%) on DRCT-2M.}
  \label{tab:drct}
  \centering
  \renewcommand{\arraystretch}{1.25}
  \scriptsize
  \setlength{\tabcolsep}{3pt}
\resizebox{\ifdim\width>\textwidth\textwidth\else\width\fi}{!}{\scalebox{\tabscale}{%
  \begin{tabular}{lccccccccccccccccc}
    \toprule\addlinespace[4pt]
    Method & SDXL-Ref & SDXL-Ctrl & LCM-SDv1.5 & LCM-SDXL & LDM & SD-Ctrl & SD-Turbo & SD2.1-Ctrl & SDXL-Turbo & SD2.1 & SDv2-DR & SDv1-DR & SDv1.4 & SDv1.5 & SDXL-DR & SDXL & \textbf{Avg.} \\
    \midrule
    UnivFD\,\raisebox{-0.25ex}{\textcolor{venuecolor}{\tiny CVPR'23}} & 54.30 & 72.80 & 53.48 & 65.20 & 81.27 & 66.26 & 55.25 & 63.97 & 52.42 & 56.08 & 52.56 & 56.47 & 55.16 & 54.78 & 52.18 & 62.04 & 59.64 \\
    FatFormer\,\raisebox{-0.25ex}{\textcolor{venuecolor}{\tiny CVPR'24}} & 48.60 & 61.06 & 48.61 & 49.97 & 52.93 & 49.97 & 48.59 & 50.20 & 48.54 & 48.56 & 53.71 & 54.98 & 48.55 & 48.55 & 51.78 & 48.55 & 50.82 \\
    NPR\,\raisebox{-0.25ex}{\textcolor{venuecolor}{\tiny CVPR'24}} & 47.72 & 57.20 & 49.16 & 47.72 & 48.85 & 48.11 & 48.28 & 48.57 & 48.07 & 48.50 & 50.78 & 52.60 & 48.84 & 48.77 & 53.03 & 47.76 & 49.62 \\
    DRCT\,\raisebox{-0.25ex}{\textcolor{venuecolor}{\tiny ICML'24}} & 85.86 & 80.19 & \underline{99.65} & 78.73 & \underline{99.90} & \underline{99.89} & 91.94 & 95.04 & 70.54 & 96.42 & 92.63 & \underline{99.88} & \underline{99.90} & \underline{99.88} & 72.27 & 83.90 & 90.41 \\
    C2P-CLIP\,\raisebox{-0.25ex}{\textcolor{venuecolor}{\tiny AAAI'25}} & 50.92 & 72.09 & 50.67 & 53.56 & 63.61 & 57.02 & 51.07 & 54.28 & 50.44 & 50.95 & 50.91 & 55.89 & 51.88 & 51.64 & 50.68 & 52.98 & 54.29 \\
    AIDE\,\raisebox{-0.25ex}{\textcolor{venuecolor}{\tiny ICLR'25}} & 66.05 & 53.42 & 69.95 & 54.13 & 60.74 & 64.80 & 53.03 & 52.95 & 53.07 & 58.09 & 50.02 & 54.52 & 74.82 & 74.28 & 50.01 & 53.11 & 58.94 \\
    AlignedForensics\,\raisebox{-0.25ex}{\textcolor{venuecolor}{\tiny ICLR'25}} & 80.49 & 84.08 & \textbf{99.85} & 89.17 & \textbf{99.95} & \textbf{99.98} & \textbf{99.88} & \underline{99.30} & 88.15 & \textbf{99.77} & \textbf{99.88} & \textbf{99.98} & \textbf{99.97} & \textbf{99.98} & 92.68 & 88.90 & 95.13 \\
    Effort\,\raisebox{-0.25ex}{\textcolor{venuecolor}{\tiny ICML'25}} & 64.81 & 65.94 & 74.96 & 69.65 & 49.99 & 68.00 & 64.20 & 62.27 & 58.91 & 84.89 & 54.37 & 64.80 & 85.73 & 85.66 & 54.10 & 70.49 & 67.42 \\
    SAFE\,\raisebox{-0.25ex}{\textcolor{venuecolor}{\tiny KDD'25}} & 50.11 & 55.28 & 50.13 & 49.94 & 50.22 & 49.96 & 49.97 & 50.04 & 49.95 & 49.97 & 50.35 & 50.22 & 50.01 & 50.02 & 50.67 & 49.92 & 50.42 \\
    DDA\,\raisebox{-0.25ex}{\textcolor{venuecolor}{\tiny NeurIPS'25}} & \underline{97.47} & \underline{99.45} & 96.98 & \underline{98.60} & 99.32 & 98.96 & 98.31 & 99.24 & \underline{95.53} & 98.54 & \underline{99.47} & 99.18 & 99.06 & 99.06 & \underline{98.11} & \underline{98.18} & \underline{98.47} \\
    GAPL\,\raisebox{-0.25ex}{\textcolor{venuecolor}{\tiny CVPR'26}} & 96.57 & 95.55 & 98.44 & 94.99 & 99.21 & 98.27 & 95.48 & 96.33 & 93.61 & 97.07 & 90.29 & 93.34 & 97.94 & 97.88 & 84.34 & 94.33 & 95.23 \\
    \midrule
    \rowcolor{oursrow} \textbf{\method{} (Ours)} & \textbf{99.21} & \textbf{99.48} & 99.38 & \textbf{99.40} & 99.41 & 99.52 & \underline{99.17} & \textbf{99.44} & \textbf{98.82} & \underline{99.15} & 99.45 & 99.48 & 99.15 & 99.10 & \textbf{98.98} & \textbf{99.16} & \textbf{99.27} \\
    \addlinespace[4pt]\bottomrule
  \end{tabular}}}
\end{table}

\clearpage
\begin{table}[!t]
  \caption{Comparison of balanced accuracy (\%) on DDA-COCO and of accuracy (\%) on the synthetic images of EvalGEN.}
  \label{tab:ddacoco_evalgen}
  \centering
  \renewcommand{\arraystretch}{1.25}
  \scriptsize
  \setlength{\tabcolsep}{3pt}
\resizebox{\ifdim\width>\textwidth\textwidth\else\width\fi}{!}{\scalebox{\tabscale}{%
  \begin{tabular}{lccccccccccccc}
    \toprule\addlinespace[4pt]
    Method & \multicolumn{7}{c}{DDA-COCO} & \multicolumn{6}{c}{EvalGEN} \\
    \cmidrule(lr){2-8} \cmidrule(lr){9-14}
    & VAE-ema & VAE-mse & SDXL & SD2.1 & SD3.5 & Flux & \textbf{Avg.} & Flux & GoT & Infinity & NOVA & OmniGen & \textbf{Avg.} \\
    \midrule
    UnivFD\,\raisebox{-0.25ex}{\textcolor{venuecolor}{\tiny CVPR'23}} & 54.40 & 53.24 & 51.52 & 53.29 & 51.28 & 50.21 & 52.32 & 3.94 & 10.27 & 15.30 & 39.72 & 7.71 & 15.39 \\
    FatFormer\,\raisebox{-0.25ex}{\textcolor{venuecolor}{\tiny CVPR'24}} & 51.49 & 53.17 & 50.77 & 53.31 & 51.63 & 49.42 & 51.63 & 1.59 & 16.14 & 19.03 & 91.85 & 7.22 & 27.17 \\
    NPR\,\raisebox{-0.25ex}{\textcolor{venuecolor}{\tiny CVPR'24}} & 49.39 & 48.88 & 48.30 & 48.85 & 48.52 & 50.46 & 49.07 & 2.04 & 1.10 & 6.01 & 2.42 & 5.57 & 3.43 \\
    DRCT\,\raisebox{-0.25ex}{\textcolor{venuecolor}{\tiny ICML'24}} & 83.31 & 76.99 & 62.49 & 76.82 & 51.47 & 50.24 & 66.89 & 36.26 & 49.59 & 63.40 & 49.68 & 55.70 & 50.93 \\
    C2P-CLIP\,\raisebox{-0.25ex}{\textcolor{venuecolor}{\tiny AAAI'25}} & 51.42 & 51.27 & 50.22 & 51.23 & 51.28 & 49.98 & 50.90 & 1.78 & 6.71 & 11.39 & 46.57 & 2.66 & 13.82 \\
    AIDE\,\raisebox{-0.25ex}{\textcolor{venuecolor}{\tiny ICLR'25}} & 50.14 & 50.12 & 50.10 & 50.09 & 50.03 & 50.04 & 50.09 & 14.28 & 18.93 & 3.20 & 12.65 & 16.38 & 13.09 \\
    AlignedForensics\,\raisebox{-0.25ex}{\textcolor{venuecolor}{\tiny ICLR'25}} & \textbf{99.80} & \textbf{99.69} & 91.98 & \textbf{99.67} & 75.61 & 51.16 & 86.32 & 20.96 & 64.47 & 71.79 & 78.48 & 71.39 & 61.42 \\
    Effort\,\raisebox{-0.25ex}{\textcolor{venuecolor}{\tiny ICML'25}} & 52.63 & 52.34 & 52.14 & 52.40 & 51.67 & 50.09 & 51.88 & 85.56 & \underline{99.17} & 92.98 & 95.67 & 98.66 & 94.41 \\
    SAFE\,\raisebox{-0.25ex}{\textcolor{venuecolor}{\tiny KDD'25}} & 50.35 & 50.39 & 50.21 & 50.42 & 50.07 & 50.25 & 50.28 & 1.00 & 0.48 & 1.88 & 0.59 & 1.56 & 1.10 \\
    DDA\,\raisebox{-0.25ex}{\textcolor{venuecolor}{\tiny NeurIPS'25}} & 99.15 & \underline{99.35} & \underline{96.99} & \underline{99.36} & \underline{83.56} & \underline{78.92} & \underline{92.89} & 88.68 & 99.14 & 97.73 & \underline{99.53} & \underline{99.20} & \underline{96.86} \\
    GAPL\,\raisebox{-0.25ex}{\textcolor{venuecolor}{\tiny CVPR'26}} & 89.60 & 88.93 & 78.36 & 88.90 & 70.95 & 58.85 & 79.27 & \underline{90.76} & 97.46 & \underline{98.39} & 97.59 & 96.31 & 96.10 \\
    \midrule
    \rowcolor{oursrow} \textbf{\method{} (Ours)} & \underline{99.28} & 99.27 & \textbf{98.13} & 99.27 & \textbf{85.17} & \textbf{85.19} & \textbf{94.39} & \textbf{96.84} & \textbf{99.64} & \textbf{99.98} & \textbf{99.60} & \textbf{99.61} & \textbf{99.14} \\
    \addlinespace[4pt]\bottomrule
  \end{tabular}}}
\end{table}

\begin{table}[!t]
  \caption{Comparison of balanced accuracy (\%) on ForenSynths.}
  \label{tab:forensynths}
  \centering
  \renewcommand{\arraystretch}{1.25}
  \scriptsize
  \setlength{\tabcolsep}{3pt}
\resizebox{\ifdim\width>\textwidth\textwidth\else\width\fi}{!}{\scalebox{\tabscale}{%
  \begin{tabular}{lcccccccccccccc}
    \toprule\addlinespace[4pt]
    Method & BigGAN & CRN & CycleGAN & DeepFake & GauGAN & IMLE & ProGAN & SAN & SITD & StarGAN & StyleGAN & StyleGAN2 & WFIR & \textbf{Avg.} \\
    \midrule
    UnivFD\,\raisebox{-0.25ex}{\textcolor{venuecolor}{\tiny CVPR'23}} & 87.50 & 55.67 & 96.92 & 69.39 & 98.79 & 68.11 & \underline{99.38} & 58.22 & 62.22 & 95.12 & 79.99 & 69.41 & 69.20 & 77.69 \\
    FatFormer\,\raisebox{-0.25ex}{\textcolor{venuecolor}{\tiny CVPR'24}} & 97.90 & 69.32 & \underline{99.32} & \underline{86.03} & 99.17 & 69.45 & 99.21 & 62.10 & 80.83 & \textbf{99.75} & 87.51 & 81.63 & 87.50 & 86.13 \\
    NPR\,\raisebox{-0.25ex}{\textcolor{venuecolor}{\tiny CVPR'24}} & 43.25 & 0.00 & 59.83 & 33.30 & 35.62 & 0.00 & 50.45 & 48.40 & 22.22 & 49.47 & 50.18 & 50.02 & 49.65 & 37.88 \\
    DRCT\,\raisebox{-0.25ex}{\textcolor{venuecolor}{\tiny ICML'24}} & 52.45 & 49.80 & 48.83 & 56.38 & 49.45 & 49.26 & 53.02 & 86.76 & 54.44 & 54.38 & 54.45 & 53.74 & 50.15 & 54.86 \\
    C2P-CLIP\,\raisebox{-0.25ex}{\textcolor{venuecolor}{\tiny AAAI'25}} & 92.07 & \underline{92.75} & \textbf{99.39} & \textbf{91.85} & 98.20 & \textbf{97.68} & 97.10 & 74.20 & \textbf{87.22} & \underline{99.70} & 76.97 & 58.51 & 65.35 & 87.00 \\
    AIDE\,\raisebox{-0.25ex}{\textcolor{venuecolor}{\tiny ICLR'25}} & 43.68 & 19.96 & 51.73 & 45.54 & 39.73 & 19.38 & 49.31 & 51.37 & 30.56 & 51.43 & 49.57 & 49.20 & 49.20 & 42.36 \\
    AlignedForensics\,\raisebox{-0.25ex}{\textcolor{venuecolor}{\tiny ICLR'25}} & 50.52 & 49.95 & 49.69 & 52.69 & 50.24 & 49.91 & 50.35 & 61.87 & 50.56 & 51.50 & 51.46 & 50.31 & 50.00 & 51.47 \\
    Effort\,\raisebox{-0.25ex}{\textcolor{venuecolor}{\tiny ICML'25}} & 73.60 & 39.20 & 81.49 & 64.59 & 62.52 & 29.54 & 70.08 & 58.90 & 83.89 & 96.02 & 56.42 & 47.67 & 70.35 & 64.17 \\
    SAFE\,\raisebox{-0.25ex}{\textcolor{venuecolor}{\tiny KDD'25}} & 48.95 & 0.11 & 52.58 & 49.09 & 42.60 & 0.10 & 49.96 & 50.46 & 35.83 & 49.95 & 49.99 & 49.99 & 49.10 & 40.67 \\
    DDA\,\raisebox{-0.25ex}{\textcolor{venuecolor}{\tiny NeurIPS'25}} & 90.97 & 86.98 & 72.47 & 76.53 & 92.74 & 89.71 & 92.80 & \underline{94.75} & 58.61 & 72.74 & 87.80 & 90.17 & 52.10 & 81.41 \\
    GAPL\,\raisebox{-0.25ex}{\textcolor{venuecolor}{\tiny CVPR'26}} & \underline{98.62} & 67.85 & 97.54 & 84.88 & \textbf{99.63} & 67.12 & \textbf{99.83} & \textbf{96.58} & 82.50 & 97.05 & \underline{97.89} & \textbf{98.62} & \textbf{94.10} & \underline{90.94} \\
    \midrule
    \rowcolor{oursrow} \textbf{\method{} (Ours)} & \textbf{99.28} & \textbf{96.07} & 97.37 & 55.26 & \underline{99.55} & \underline{96.08} & 98.99 & 88.81 & \underline{84.72} & 99.25 & \textbf{98.21} & \underline{98.19} & \underline{91.65} & \textbf{92.57} \\
    \addlinespace[4pt]\bottomrule
  \end{tabular}}}
\end{table}

\begin{table}[!t]
  \caption{Comparison of balanced accuracy (\%) on the ManipulationBench subset of DailyBench; rows other than ours are as reported by DailyBench, including its own detector FPD.}
  \label{tab:dailybench}
  \centering
  \renewcommand{\arraystretch}{1.25}
  \scriptsize
  \setlength{\tabcolsep}{3pt}
\resizebox{\ifdim\width>\textwidth\textwidth\else\width\fi}{!}{\scalebox{\tabscale}{%
  \begin{tabular}{lcccccccc}
    \toprule\addlinespace[4pt]
    Method & \begin{tabular}[c]{@{}c@{}}FLUX-Fill\\(rand.)\end{tabular} & \begin{tabular}[c]{@{}c@{}}FLUX-Fill\\(obj.)\end{tabular} & \begin{tabular}[c]{@{}c@{}}FLUX.2\\-klein\end{tabular} & \begin{tabular}[c]{@{}c@{}}Qwen-\\Edit\end{tabular} & \begin{tabular}[c]{@{}c@{}}Step1X-\\Edit\end{tabular} & \begin{tabular}[c]{@{}c@{}}Nano\\Banana~2\end{tabular} & \begin{tabular}[c]{@{}c@{}}GPT-\\Image~2\end{tabular} & \textbf{Avg.} \\
    \midrule
    UnivFD\,\raisebox{-0.25ex}{\textcolor{venuecolor}{\tiny CVPR'23}} & 50.16 & 50.83 & 50.72 & 50.39 & 50.72 & 50.26 & 50.23 & 50.47 \\
    FatFormer\,\raisebox{-0.25ex}{\textcolor{venuecolor}{\tiny CVPR'24}} & 49.86 & 49.95 & 50.11 & 50.05 & 50.26 & 50.00 & 56.96 & 51.02 \\
    NPR\,\raisebox{-0.25ex}{\textcolor{venuecolor}{\tiny CVPR'24}} & 45.09 & 46.02 & 49.81 & 55.17 & 48.77 & 55.53 & 80.33 & 54.38 \\
    DRCT\,\raisebox{-0.25ex}{\textcolor{venuecolor}{\tiny ICML'24}} & 51.00 & 50.86 & 50.56 & 52.17 & 50.13 & 57.21 & 63.99 & 53.70 \\
    AIDE\,\raisebox{-0.25ex}{\textcolor{venuecolor}{\tiny ICLR'25}} & 51.13 & 50.65 & 49.63 & 58.63 & 57.54 & 54.43 & 60.43 & 54.63 \\
    AlignedForensics\,\raisebox{-0.25ex}{\textcolor{venuecolor}{\tiny ICLR'25}} & 50.51 & 50.44 & 50.42 & 73.18 & 50.43 & \textbf{64.14} & 56.96 & 56.58 \\
    Effort\,\raisebox{-0.25ex}{\textcolor{venuecolor}{\tiny ICML'25}} & 45.60 & 45.70 & 50.67 & 56.98 & 53.52 & 60.85 & 68.99 & 54.61 \\
    SAFE\,\raisebox{-0.25ex}{\textcolor{venuecolor}{\tiny KDD'25}} & 52.22 & 51.94 & 49.99 & 48.96 & 48.26 & 49.46 & \textbf{86.67} & 55.35 \\
    DDA\,\raisebox{-0.25ex}{\textcolor{venuecolor}{\tiny NeurIPS'25}} & \underline{54.54} & \underline{54.82} & \underline{62.29} & \textbf{89.20} & \textbf{62.88} & \underline{62.48} & 81.96 & \underline{66.88} \\
    FPD\,\raisebox{-0.25ex}{\textcolor{venuecolor}{\tiny arXiv'26}} & 53.68 & 54.09 & 58.00 & 80.47 & \underline{60.70} & 53.22 & 83.71 & 63.41 \\
    \midrule
    \rowcolor{oursrow} \textbf{\method{} (Ours)} & \textbf{62.58} & \textbf{65.89} & \textbf{64.86} & \underline{81.18} & 58.12 & 59.94 & \underline{85.69} & \textbf{68.32} \\
    \addlinespace[4pt]\bottomrule
  \end{tabular}}}
\end{table}

\section{More Ablation Studies}
\label{app:ablation}

\paragraph{Full ablation results.}
Table~\ref{tab:ablation_full} lists the numbers behind Figure~\ref{fig:ablation}.

\begin{table}[!htb]
  \caption{\textbf{Ablation.} Balanced accuracy (\%) of the variants of Figure~\ref{fig:ablation} on GenImage, AIGCDetect, Chameleon and WildRF. All rows use the same DINOv3 backbone with LoRA and the same aligned training pairs. Top: three ways of using the aligned pairs; bottom: \method{} with one objective removed.}
  \label{tab:ablation_full}
  \centering
  \renewcommand{\arraystretch}{1.25}
  \scriptsize
  \setlength{\tabcolsep}{3pt}
\resizebox{\ifdim\width>\textwidth\textwidth\else\width\fi}{!}{\scalebox{\tabscale}{%
  \begin{tabular}{lccccc}
    \toprule\addlinespace[4pt]
    Method & GenImage & AIGCDetect & Chameleon & WildRF & \textbf{Avg.} \\
    \midrule
    Image-level BCE on the aligned pairs & 94.58 & 96.06 & 85.89 & 94.06 & 92.65 \\
    \ \ + mixed views, image-level BCE & \underline{97.33} & \underline{97.38} & 82.84 & 93.17 & 92.68 \\
    \ \ + mixed views, per-patch label & \textbf{97.45} & 96.92 & 79.29 & 93.62 & 91.82 \\
    \rowcolor{oursrow} \textbf{\method{} (Ours)} & 96.83 & \textbf{97.75} & \textbf{90.01} & \textbf{97.59} & \textbf{95.55} \\
    \midrule
    w/o $\mathcal{L}_{\mathrm{rel}}$ & 96.32 & 96.77 & 83.20 & 95.82 & 93.03 \\
    w/o $\mathcal{L}_{\mathrm{coh}}$ & 96.05 & 96.43 & \underline{86.83} & 96.89 & 94.05 \\
    w/o $\mathcal{L}_{\mathrm{rank}}$ & 95.72 & 97.05 & 86.61 & \underline{97.20} & \underline{94.14} \\
    \addlinespace[4pt]\bottomrule
  \end{tabular}}}
\end{table}

\paragraph{DDA on the same backbone.}
Table~\ref{tab:dda_backbone} compares DDA with its released DINOv2 backbone, DDA adapted to our DINOv3 backbone with the same LoRA, and \method{} on the eleven benchmarks (Section~\ref{sec:ablation}).

\begin{table}[!htb]
  \caption{\textbf{DDA and \method{} on the same backbone.} Balanced accuracy (\%) on the eleven benchmarks of Table~\ref{tab:main} for the released DDA model (DINOv2), for DDA adapted from its official code to our DINOv3 backbone with LoRA, and for \method{}, which uses the same backbone, adaptation and aligned pairs.}
  \label{tab:dda_backbone}
  \centering
  \renewcommand{\arraystretch}{1.25}
  \scriptsize
  \setlength{\tabcolsep}{2.5pt}
\resizebox{\ifdim\width>\textwidth\textwidth\else\width\fi}{!}{\scalebox{\tabscale}{%
  \begin{tabular}{lcccccccccccc}
    \toprule\addlinespace[4pt]
    \multirow{2}{*}{Method} & \multicolumn{8}{c}{Standard Benchmarks} & \multicolumn{3}{c}{In-the-Wild Benchmarks} & \\
    \cmidrule(lr){2-9} \cmidrule(lr){10-12} \cmidrule(lr){13-13}
     & GenImage & AIGCDetect & DRCT-2M & DDA-COCO & EvalGEN & Synthbuster & ForenSynths & UnivFD & Chameleon & SynthWildX & WildRF & \textbf{Avg.} \\
    \midrule
    DDA, released model (DINOv2) & 91.74 & 87.83 & 98.47 & 92.89 & 96.86 & 85.52 & 81.41 & 82.58 & 82.41 & 90.87 & 90.20 & 89.16 \\
    DDA on DINOv3 & \textbf{97.93} & 96.09 & 99.04 & \textbf{95.98} & \textbf{99.40} & 88.77 & 87.50 & 85.52 & \textbf{91.82} & 89.01 & 94.17 & 93.20 \\
    \midrule
    \rowcolor{oursrow} \textbf{\method{} (Ours)} & 96.83 & \textbf{97.75} & \textbf{99.27} & 94.39 & 99.14 & \textbf{97.00} & \textbf{92.57} & \textbf{94.22} & 90.01 & \textbf{93.60} & \textbf{97.59} & \textbf{95.67} \\
    \addlinespace[4pt]\bottomrule
  \end{tabular}}}
\end{table}

\begin{figure}[!htb]
  \centering
  \includegraphics[width=\textwidth]{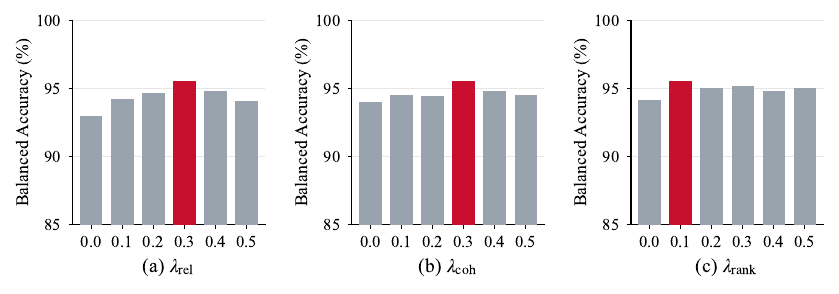}
  \caption{\textbf{Sensitivity to the loss weights.} Mean balanced accuracy (\%) over GenImage, AIGCDetect, Chameleon and WildRF when one weight varies and the other two keep their default values; the red bar is the default setting.}
  \label{fig:sensitivity}
\end{figure}

\paragraph{Sensitivity to the loss weights.}
Figure~\ref{fig:sensitivity} varies one loss weight at a time on the same four benchmarks, with the other two kept at their default values $(\lambda_{\mathrm{rel}},\lambda_{\mathrm{coh}},\lambda_{\mathrm{rank}})=(0.3,\,0.3,\,0.1)$. Setting a weight to zero reproduces the corresponding row of Table~\ref{tab:ablation_full}. \method{} is robust to the choice of loss weights. The default is the best point on every axis, every nonzero weight in $[0.1, 0.5]$ stays within $1.5\%$ of it, and every setting remains above the image-level label.

\end{document}